\def\submitversion{2}%
\def\fullversion{2}%
\if\submitversion1
\documentclass[conference]{IEEEtran}
\usepackage[subtle,lists=tight]{savetrees}
\fi
\if\submitversion2
\documentclass[onecolumn,draftclsnofoot,12pt]{IEEEtran}
\fi
\usepackage[table]{xcolor}
\usepackage[cmex10]{amsmath}
\usepackage{amssymb}
\usepackage{amsthm}
\usepackage{mathtools}
\usepackage{booktabs}
\usepackage{bm}
\usepackage{graphicx}
\IEEEoverridecommandlockouts
\usepackage[square,sort,comma,numbers]{natbib}
\usepackage{algorithm}
\usepackage{algorithmic}
\usepackage{balance}
\usepackage{multirow}
\usepackage{stmaryrd}
\newtheorem{theorem}{Theorem}
\newtheorem{remark}{Remark}
 
 \newtheorem{proposition}{Proposition}
 
 \newtheorem{lemma}{Lemma}
 
\theoremstyle{definition}
 \newtheorem{definition}{Definition}

\theoremstyle{remark}
\newcommand{\Proj}{\text{Proj}}

\newcommand{\diff}{\mathrm{d}}

\DeclareMathOperator*{\argmin}{arg\,min}

\newcommand{\cP}{\mathcal{P}}
\newcommand{\cN}{\mathcal{N}}
\newcommand{\bR}{\mathbb{R}}
\newcommand{\bF}{\mathbb{F}}
\newcommand{\bQ}{\mathbb{Q}}

\newcommand{\cW}{\mathcal{W}}
\newcommand{\Reg}{\varepsilon}
\newcommand{\orho}{\bar{\rho}}
\newcommand{\hP}{\hat{\mathbb{P}}}
\newcommand{\bP}{\mathbb{P}}

\DeclarePairedDelimiterX{\inp}[2]{\langle}{\rangle}{#1, #2}

\title{A Data-Driven Approach to Robust Hypothesis Testing Using Sinkhorn Uncertainty Sets}
\author{Jie Wang and Yao Xie
\thanks{J.~Wang and Y. Xie are with H. Milton Stewart School of Industrial and Systems Engineering, Georgia Institute of Technology.
Email: jwang3163@gatech.edu, yao.xie@isye.gatech.edu.
The project is funded by an NSF CAREER Award CCF-1650913, and DMS-2134037, DMS-1938106, and DMS-1830210.
}
}

\begin{document}

\maketitle
\begin{abstract}
Hypothesis testing for small-sample scenarios is a practically important problem.
In this paper, we investigate the robust hypothesis testing problem in a data-driven manner, where we seek the worst-case detector over distributional uncertainty sets centered around the empirical distribution from samples using Sinkhorn distance. 
Compared with the Wasserstein robust test, the corresponding least favorable distributions are supported beyond the training samples, which provides a more flexible detector.
Various numerical experiments are conducted on both synthetic and real datasets to validate the competitive performances of our proposed method. 
\end{abstract}
\section{Introduction}
As a fundamental problem in statistics, hypothesis testing plays a key role in general scientific discovery areas such as anomaly detection and model criticism. 
The goal of hypothesis testing is to determine which one among given hypotheses is true within a certain error probability level.
Unfortunately, the data-generating distributions are usually unknown so that it is difficult to obtain the optimal test leveraging the Neyman-Pearson Lemma~\citep{neyman1933ix}.
Although training samples from target distributions are often available, we cannot obtain reliable estimates of the underlying distributions for small-sample cases.
Therefore, hypothesis testing for small-sample scenarios is a challenging task, and it commonly arises in many practical applications such as health care~\citep{Schober19}, anomaly detection~\citep{Chandola2009, savage2014anomaly, ahmed2016survey}, and change-point detection~\citep{Vincent08,xie2020sequential,Liyanchange_21,xie2022minimaxdec}.

Various robust detectors are developed in existing literature to capture the distributional uncertainty such as distribution mis-specification and adversarial data perturbation.
They are constructed by seeking the worst-case detectors over distributional uncertainty sets that contain candidate distributions under the null and alternative hypotheses.
The earliest work on robust detectors dates back to Huber's masterpiece~\citep{Peter65}, which constructs the uncertainty sets as probability balls centered around nominal distributions using total-variation distance.
However, it is computationally intractable to obtain the corresponding optimal tests, especially for multivariate settings.
Recent works~\citep{Levy09, gul2017minimax} construct the uncertainty sets as balls using KL-divergence centered around empirical distributions such that all distributions within the sets are supported only on training samples.
We remark that for small-sample scenarios, this choice is too restrictive since there is a non-negligible probability that new samples are outside the support of training samples.

We consider a data-driven robust hypothesis testing problem when the sample size is small.
A closely related work~\citep{gao18robust} constructs the distributional uncertainty sets using Wasserstein distance.
The Wasserstein distance takes account into the geometry of sample space and therefore is suitable for comparing distributions with non-overlapping supports, and hedging against data outliers~\citep{gao2016distributionally}.
However, the Wasserstein robust test is not without limitation.
As shown in \citep{xie2021robust}, the induced optimal test is a likelihood ratio test between \emph{least favorable distributions}~(LFDs) supported on training samples, which may not be applicable if testing samples do not have the same support as those training samples.
Although it is possible to extend LFDs into the whole sample space using kernel smoothing~\citep{xie2021robust} or $k$-nearest neighbors~\citep{wang2021classconditioned} algorithms, the corresponding test may not achieve good performances as the distributional estimates are not necessarily reliable.
References~\citep{Zhongchang21,sun2022kernel} address the drawback of Wasserstein distance by constructing uncertainty sets using maximum mean discrepancy~(MMD). 
To maintain computational tractability, their goal is to find the optimal detector so that asymptotically the type-II error exponent is maximized and the type-I error is below a threshold.
However, this test may not be optimal in small-sample cases and as demonstrated in some numerical experiments (see Section~\ref{Sec:application}), the MMD robust test may not achieve the best performances.

In this paper, we develop a new robust testing framework leveraging the idea of distributionally robust optimization~(DRO) with Sinkhorn distance~\citep{wang2021sinkhorn}, which, as a variant of Wasserstein distance with stochastic transport mapping, is defined as the cheapest transport cost between two distributions with entropic regularization~\citep{cuturi2013sinkhorn}.
Specifically, we study the robust hypothesis testing problem by seeking the worst-case detector over ambiguity sets so that the risk is minimized, where the ambiguity sets are constructed using Sinkhorn distance centered around the empirical distributions from samples.
The resulted worst-case detector is well-defined for samples outside the training samples, which usually leads to better generalization performances than the previous framework.
Our contributions are summarized as follows.
\begin{enumerate}
    \item 
We formulate the problem of robust hypothesis testing as an infinite-dimensional optimization that seeks the optimal detector and LFDs jointly, which is challenging to solve in general. 
We derive its dual reformulation leveraging tools from distributionally robust optimization, which enables us to derive the optimal detector in two steps:
\begin{enumerate}
    \item[(I)]
    Given a fixed pair of distributions, we first find the corresponding optimal detector.
    \item[(II)]
Then we find the LFDs by solving an infinite-dimensional convex optimization. We leverage the Monte-Carlo approximation idea to solve a finite-dimensional problem instead.
\end{enumerate}
\item
Various numerical experiments using both synthetic and real datasets are conducted to demonstrate competitive performances of our proposed method.
\end{enumerate}

The rest of this paper is organized as follows.
Section~\ref{Section:setup} describes the main formulation and a brief introduction to Sinkhorn DRO,
Section~\ref{Sec:methodology} develops the methodology for solving the robust hypothesis testing problem,
Section~\ref{Sec:application} reports several numerical results,
and Section~\ref{Sec:conclusion} provides some concluding remarks.
\if\fullversion1 All omitted proofs and experiment details can be found in \citep{jie22isit_full}.
\fi
\if\fullversion2
All omitted proofs and other details can be found in Appendix.
\fi

\textit{Notations:} 
Denote $\bF$ as the set $\{0,1\}$.
The base of the logarithm function $\log$ is $e$.
For any non-negative integer $N$, define $[N] := \{1, \ldots, N\}$.
Given a reference measure $\nu$ supported on $\Omega$ and a function $f:~\Omega\to\bR$, define the essential supremum $\text{ess-sup}~f=\inf\{t:~\nu\{f(z)>t\}=0\}$.
We write $\bP\ll\nu$ if the distribution $\bP$ is absolutely continuous with respect to the measure $\nu$. 
Denote by $\text{supp}(\bP)$ the support of the distribution $\bP$.

\section{Problem Setup}\label{Section:setup}
Let $\Omega\subseteq\mathbb{R}^d$ be the sample space in which the observed samples take their values, and $\cP(\Omega)$ be the set of all distributions supported on $\Omega$.
Denote by $\cP_0, \cP_1\subseteq\cP(\Omega)$ the uncertainty sets under hypotheses $H_0$ and $H_1$, respectively. 
Given two sets of training samples $\{x_1^k,\ldots,x_{n_k}^k\}$ generated from $\bP_k\in\cP_k$ for $k\in\bF$, denote the corresponding empirical distributions as $\hP_k=\frac{1}{n_k}\sum_{i=1}^{n_k}\delta_{x_i^k}$.
For notation simplicity, assume that $n_0=n_1=n$, but our formulation can be naturally extended for unequal sample sizes.
Given a new testing sample $\omega$, the goal of \emph{composite hypothesis testing} is to distinguish between the null hypothesis $H_0:~\omega\sim \bP_0$ and the alternative hypothesis $H_1:~\omega\sim \bP_1$, where $\bP_k\in\cP_k$ for $k\in\bF$.
For a detector $T:~\Omega\to\bR$, it accepts the null hypothesis $H_0$ when $T(\omega)\ge0$ and otherwise it accepts the alternative hypothesis $H_1$.
Under the Bayesian setting, the risk of this detector is quantified as the summation of type-I and type-II error:
\[
\mathcal{R}(T;\bP_0,\bP_1)=\bP_0\{\omega: T(\omega)<0\} + \bP_1\{\omega: T(\omega)\ge0\}.
\]
Since the objective function is highly non-convex, we replace it with its tight upper bound via convex approximations of the indicator function as discovered in \citep{nemirovski2007convex,xie2021robust}:
\[
\Phi(T;\bP_0,\bP_1) = \mathbb{E}_{\bP_0}[\ell\circ(-T)(\omega)] + \mathbb{E}_{\bP_1}[\ell\circ T(\omega)],
\]
\if\submitversion1
where $\ell$ is a \emph{generating function} (see Definition~\ref{Def:generating}).
\fi
\if\submitversion2
where $\ell$ is a \emph{generating function} (see Definition~\ref{Def:generating}) so that it always holds that
\[\Phi(T;\bP_0,\bP_1)\ge \mathcal{R}(T;\bP_0,\bP_1).
\]
\fi

\begin{definition}[Generating Function]\label{Def:generating}
A generating function $\ell:~\bR\to\bR_+\cup\{\infty\}$ is a non-negative valued, non-decreasing, convex function so that $\ell(0)=1$ and $\lim_{t\to-\infty}\ell(t)=0$.
\end{definition}
Table~\ref{tab:generating:function} lists some common choices of generating function $\ell$ and the corresponding optimal detector, in which the first, second, fourth one has been considered in existing literature \citep{goldenshluger2015hypothesis}, \citep{cheng2020classification}, \citep{xie2021robust}, respectively.
In this paper, we develop a minimax test that optimizes the worst-case risk function over all distributions within ambiguity sets $\cP_0$ and $\cP_1$:
\begin{equation}
\inf_{T}\sup_{\bP_k\in\cP_k, {k\in\bF}}~\Phi(T;\bP_0,\bP_1),\label{Eq:minimax:test}
\end{equation}
where the sets $\cP_k, {k\in\bF}$ are formulated using Sinkhorn distance:
\begin{equation}\label{Eq:Sinkhorn:ambiguity}
\cP_k = \{\bP_k\in \cP:~\cW_{\Reg}(\hP_k, \bP_k)\le \rho_k\}.
\end{equation}
The resulting worst-case distributions $\bP_k^*, {k\in\bF}$ in \eqref{Eq:minimax:test} are called the \emph{least favorable distributions}~(LFDs) in literature.
Leverging results from \citep[Theorem~1]{xie2021robust}, we can argue that the approximation \eqref{Eq:minimax:test} is near optimal for developing the robust test to optimize $\mathcal{R}(T;\bP_0,\bP_1)$, the summation of type-I and type-II error.

\begin{remark}[Batched Testing]
When given a batch of $n_{\text{Te}}$ testing samples $\omega_1,\ldots,\omega_{n_{\text{Te}}}$ generated from the same distribution and a detector $T:~\Omega\to\bR$, the decision is made based on the principle of majority vote, i.e., we accept the null hypothesis $H_0$ if 
\[
{T}(\omega_1,\ldots,\omega_{n_{\text{Te}}}):=\frac{1}{n_{\text{Te}}}\sum_{i=1}^{n_{\text{Te}}}{T}(\omega_i)<0.
\]
As shown in \citep[Proposition~1]{xie2021robust}, both type-I and type-II error for batched testing procedure decrease exponentially fast to zero as the testing sample size $n_{\text{Te}}$ increases.
\end{remark}

\subsection{Preliminaries about Sinkhorn DRO}\label{Sec:preliminary:SDRO}
In the following we review some details about Sinkhorn DRO. 
The Sinkhorn distance is a variant of the Wasserstein distance based on entropic regularization.
\begin{definition}[Sinkhorn Distance]\label{Def:Sinkhorn}
Consider any two distributions $\bP,\bQ\in \cP(\Omega)$ and let $\nu\in\mathcal{M}(\Omega)$ be a reference measure such that $\bQ\ll \nu$.
For regularization parameter $\Reg>0$, define the Sinkhorn distance between two distributions $\bP$ and $\bQ$ as
\[
\cW_{\Reg}(\bP,\bQ)=\inf_{\gamma\in\Gamma(\bP,\bQ)}~\left\{\mathbb{E}_{(x,y)\sim\gamma}\left[c(x,y)\right] +\Reg H(\gamma\|\bP\otimes\nu)\right\},
\]
where $\Gamma(\bP,\bQ)$ denotes the set of joint distributions whose first and second marginal distributions are $\bP$ and $\bQ$, respectively, $c(x,y)$ stands for the cost function, and $H(\gamma\|\bP\otimes\nu)$ denotes the relative entropy between the distribution $\gamma$ and the measure $\bP\otimes\nu$:
\[
H(\gamma\|\bP\otimes\nu) = \int \log\left(\frac{\diff \gamma(x,y)}{\diff \bP(x)\diff\nu(y)}\right)\diff\gamma(x,y).
\]
\end{definition}

With a measurable variable $f:~\Omega\to\bR$, we associate value
\begin{equation}\label{Eq:primal:Sinkhorn}
V = \sup_{\bP\in\cP}~\mathbb{E}_{\bP}[f].
\end{equation}
We construct the ambiguity set $\cP$ using Sinkhorn distance, i.e., $\cP=\{\bP\in\cP(\Omega):~\cW_{\Reg}(\hP,\bP)\le \rho\}$ for some nominal distribution $\hP$.
For instance, the nominal distribution $\hP$ can be an empirical distribution from samples.
Define the dual problem of \eqref{Eq:primal:Sinkhorn} as 
\begin{equation}\label{Eq:dual:Sinkhorn}
V_D = \inf_{\lambda\ge0}~\lambda\bar{\rho} + \lambda\Reg\int \log\left(
\mathbb{E}_{\bQ_{x,\Reg}}\left[
e^{f(z)/(\lambda\Reg)}
\right]
\right)\diff\hP(x),
\end{equation}
where we define the constant
\[
\bar{\rho} = \rho + \Reg\int \log\left(\int e^{-c(x,z)/\Reg}\diff\nu(z)\right)\diff\hP(x)
\]
and the kernel probability distribution $\bQ_{x,\Reg}$ as 
\[
\diff \bQ_{x,\Reg}(z) = \frac{e^{-c(x,z)/\Reg}}{\int e^{-c(x,u)/\Reg}\diff\nu(u)}\diff\nu(z).
\]
The distribution $\bQ_{x,\Reg}$ can be viewed as a posterior distribution of the random variable $Z$ given $X=x$, in which the prior distribution of $Z$ is proportional to $\nu$, and the likelihood model $P(X=x\mid Z=z)\propto e^{-c(x,z)/\Reg}$.
A strong duality result for the problem \eqref{Eq:primal:Sinkhorn} is provided in Theorem~\ref{Theorem:Sinkhorn} to obtain a more tractable form.
\begin{theorem}[Reformulation of Sinkhorn DRO]\label{Theorem:Sinkhorn}
Assume that
\begin{enumerate}
\item
$\nu\{z:~0\le c(x,z)<\infty\}=1$ for $\hP$-almost every $x$;
\item
$\int e^{-c(x,z)/\Reg}\diff\nu(z)<\infty$ for $\hP$-almost every $x$;
\item
$\bar{\rho}\ge0$.
\end{enumerate}
Then it holds that $V = V_D$.
Additionally, when
\begin{equation}
\begin{aligned}
&\bar{\rho}':=\bar{\rho} + \Reg\int
\log\left( 
\mathbb{E}_{\bQ_{x,\Reg}}[1_A]
\right)\diff\hP(x)<0,
\end{aligned}
\end{equation}
where the set $A:=\{z:~f(z)=\text{ess-sup}_{\nu}~f\}$, the constraint set $\cP$ for problem \eqref{Eq:primal:Sinkhorn} is active and the worst case distribution $\bP^*$ can be expressed as
\begin{equation}\label{Eq:worst:P:*}
\diff \bP^*(z) = \int \left( 
\frac{e^{f(z)/(\lambda^*\Reg)}\diff\bQ_{x,\Reg}(z)}{\mathbb{E}_{\bQ_{x,\Reg}}[e^{f(z)/(\lambda^*\Reg)}]}
\right)\diff\hP(x),
\end{equation}
where $\lambda^*>0$ is the optimal solution for the problem~\eqref{Eq:dual:Sinkhorn}.
\end{theorem}
The finite-dimensional convex problem \eqref{Eq:dual:Sinkhorn} can be efficiently solved based on bisection search with Monte-Carlo sampling on the kernel distribution $\bQ_{x,\Reg}$.
In particular, the generation of samples from $\bQ_{x,\Reg}$ is easy for many cases.
For example, when the cost function $c(x,y)=\frac{1}{2}\|x-y\|^2$ and $\nu$ is the Lebesgue measure in $\bR^d$, it holds that $\bQ_{x,\Reg}=\cN(x, \Reg I_d)$.
When the explicit density form of $\bQ_{x,\Reg}$ is not available, we can also finish this task using the acceptance-rejection method~\citep{asmussen2007stochastic}.

From the expression \eqref{Eq:worst:P:*}, we realize the regularization parameter $\Reg$ quantifies the smoothness of the worst-case distribution $\bP^*$.
Specifically, when the optimal Lagrangian multiplier $\lambda^*>0$, the worst-case distribution maps each $x\in\text{supp}(\hP)$ to a distribution whose density function with respect to $\nu$ at $z$ is proportional to $\exp\left(\frac{1}{\Reg}(f(z) - \lambda^*c(x,z))\right)$.
When $\Reg\to0$, the distribution $\bP^*$ is discrete and one recovers the classical Wasserstein DRO formulation.
When $\Reg\to\infty$, each sample is moved uniformly so that the distribution $\bP^*$ is a uniform measure with respect to $\nu$. 
See \citep{wang2021sinkhorn} for a detailed discussion.

\if\submitversion1
\begin{table*}[tb]
  \centering
    \caption{Common choices of generating function, together with its corresponding optimal detector and detector risk function.}
    \label{tab:generating:function}
    \rowcolors{2}{white}{gray!30}
      \begin{tabular}{p{3cm}p{2.8cm}p{4.2cm}p{2.8cm}p{2.8cm}}
        \toprule
        $\ell(t)$ & $T^*$ &    $\psi(r)$ &  $1 - \Phi^*(\bP_0,\bP_1)/2$ \\
        \midrule
        $\exp(t)$ & $\log\sqrt{\diff\bP_0/\diff\bP_1}$ & $2\sqrt{r(1-r)}$&   $H^2(\bP_0,\bP_1)$ \\
        $\log(1+\exp(t))/\log 2$ & $\log(\diff\bP_0/\diff\bP_1)$ & $-(r\log r + (1-r)\log(1-r))/\log 2$&  $\text{JS}(\bP_0,\bP_1)/\log 2$  \\
        $(t+1)_+^2$ & $1 - 2(\diff\bP_0/\diff(\bP_0+\bP_1))$ & $4r(1-r)$ & $\chi^2(\bP_0,\bP_1)$   \\
        $(t+1)_+$ & $\text{sign}(\diff\bP_0-\diff\bP_1)$ & $2\min(r,1-r)$&  $\text{TV}(\bP_0,\bP_1)$\\
        \bottomrule
      \end{tabular}
\end{table*}
\fi 
\if\submitversion2
\begin{table*}[tb]
  \centering
    \caption{Common choices of generating function, together with its corresponding optimal detector and detector risk function.}
    \label{tab:generating:function}
    \rowcolors{2}{white}{gray!30}
      \begin{tabular}{p{3cm}p{3cm}p{3cm}p{3cm}p{3cm}}
        \toprule
        $\ell(t)$ & $T^*$ &    $\psi(r)$ &  $1 - \Phi^*(\bP_0,\bP_1)/2$ \\
        \midrule
        $\exp(t)$ & $\log\sqrt{\diff\bP_0/\diff\bP_1}$ & $2\sqrt{r(1-r)}$&   $H^2(\bP_0,\bP_1)$ \\
        $\log(1+\exp(t))/\log 2$ & $\log(\diff\bP_0/\diff\bP_1)$ & $H_2(r)/\log 2$&  $\text{JS}(\bP_0,\bP_1)/\log 2$  \\
        $(t+1)_+^2$ & $1 - 2(\diff\bP_0/\diff(\bP_0+\bP_1))$ & $4r(1-r)$ & $\chi^2(\bP_0,\bP_1)$   \\
        $(t+1)_+$ & $\text{sign}(\diff\bP_0-\diff\bP_1)$ & $2\min(r,1-r)$&  $\text{TV}(\bP_0,\bP_1)$\\
        \bottomrule
      \end{tabular}
\end{table*}
\fi 

\section{Methodology}\label{Sec:methodology}

\begin{center}
\begin{algorithm}[!t]
\caption{
Algorithm for Sinkhorn Robust Detector
} 
\label{Alg:Sinkhorn}
\begin{algorithmic}[1]\label{Alg:permutation:test}
\REQUIRE{Training samples $\{x_i^k\}_{i\in[n],k\in\bF}$ and a testing sample $\omega$.}
\FOR{$i=1,2,\ldots,n$} 
\STATE{Generate $m$ samples from $\bQ_{i,\Reg}^k, k\in\bF$ defined in \eqref{Eq:Q:i:Reg:k} and construct the corresponding empirical distribution $\hat{\bQ}_{i,\Reg}^k$.}
\STATE{$\hat{\mathbb{G}}_{i,\Reg} \leftarrow (\hat{\bQ}_{i,\Reg}^0 + \hat{\bQ}_{i,\Reg}^1)/2$.}
\STATE{Calculate weighted importance ratio function $r_{i,\Reg}^k$ valued on $\text{supp}(\hat{\mathbb{G}}_{i,\Reg})$ for $k\in\bF$.}
\ENDFOR
\STATE{Obtain $\{\ell_{i,k}\}_{i\in[n],k\in\bF}$ as the optimal solution to problem \eqref{Eq:Psi*:approximated}.}
\STATE{Recover LFDs $\bP_k^*, k\in\bF$ according to \eqref{Eq:Psi*:approximated:d}.}\\
\textbf{Return} the $k$-NN detector valued on $\omega$ according to Remark~\ref{Remark:kNN:detector}.
\end{algorithmic}
\end{algorithm}
\end{center}

In this section, we first develop a strong duality theorem to reformulate the problem \eqref{Eq:minimax:test}, then we leverage the idea of Monte-Carlo approximation to solve the reformulated problem, from which we can obtain the robust detector.
The overall procedure is summarized in Algorithm~\ref{Alg:Sinkhorn}.

\subsection{Step 1: Exchange of Infimum and Supremum}

Similar to the discussion in Section~\ref{Sec:preliminary:SDRO}, for $k\in\bF$, we define the constant 
\begin{equation}
\orho_k=\rho_k + \Reg\int \log\left(\int e^{-c(x,z)/\Reg}\diff\nu(z)\right)\diff\hP_k(x),    \label{Eq:orho:k}
\end{equation}
where $\rho_k$ is introduced in \eqref{Eq:Sinkhorn:ambiguity} to quantify the size of the Sinkhorn ambiguity set. In addition, we define kernel probability distribution $\bQ_{i,\Reg}^k$ as
\begin{equation}
\diff\bQ_{i,\Reg}^k(z)=\frac{e^{-c(x_i^k,z)/\Reg}}{\int e^{-c(x_i^k,u)/\Reg}\diff\nu(u)}\diff\nu(z).\label{Eq:Q:i:Reg:k}
\end{equation}
Proposition~\ref{Strong:Duality:inf:sup} presents our strong duality theorem, which enables us to switch the $\inf$ and $\sup$ operators in \eqref{Eq:minimax:test}.
It reveals that a robust detector can be obtained by finding the optimal detector for fixed distributions $\bP_k, k\in\bF$, and then finding the LFDs to maximize the risk over those detectors.
An expression of the optimal detector for fixed distributions is provided in Lemma~\ref{Lemma:optimal:detector}.
\begin{proposition}[Strong Duality]\label{Strong:Duality:inf:sup}
Assume that for $x\in\{x_i^k\}_{i\in[n],k\in\bF}$, it holds that $\nu\{z:~0\le c(x,z)<\infty\}=1$ and $\int e^{-c(x,z)/\Reg}\diff\nu(z)<\infty$.
When $\orho_k\ge0$ for $k\in\bF$, it holds that 
\begin{equation}
    \inf_{T:~\Omega\to\bR}~\sup_{\substack{\bP_k\in\cP_k,\\{k\in\bF}}}~\Phi(T;\bP_0,\bP_1)
=
\sup_{\substack{\bP_k\in\cP_k,\\{k\in\bF}}}\Phi^*(\bP_0,\bP_1),\label{Eq:relation:strong:duality}
\end{equation}%
where $\Phi^*(\bP_0,\bP_1)$ is the infimum of $\Phi(T;\bP_0,\bP_1)$ over all detectors $T:~\Omega\to\bR$.
\end{proposition}

\begin{lemma}[Optimal Detector~{\citep[Theorem 2]{gao18robust}}]\label{Lemma:optimal:detector}
For fixed $\bP_k, k\in\bF$, it holds that
\[
\Phi^*(\bP_0,\bP_1):= \int 
[\psi\circ r(\omega)]
\diff(\bP_0+\bP_1)(\omega),
\]
where the ratio
\begin{equation}\label{Eq:ratio:r}
r(\omega) = \frac{\diff\bP_0}{\diff(\bP_0+\bP_1)}(\omega),
\end{equation}
and
\[
\psi(r) := \min_{t\in\bR}~\{
\psi_t(r)\triangleq
(1-r)\ell(t) + r\ell(-t)\},\quad r\in[0,1].
\]
An optimal detector for $\Phi^*(\bP_0,\bP_1)$ is $T^*(\omega)=-t^*(\omega)$, where
\[
t^*(\omega):=\argmin_{t\in\bR}~\{(1-r(\omega))\ell(t) + r(\omega)\ell(-t)\}. %
\]
\end{lemma}

\begin{IEEEproof}[Proof Sketch of Proposition~\ref{Strong:Duality:inf:sup}]
The idea to show the strong duality result is as follows.
We first reformulate the infimum of $\Phi$ among all detectors (see Lemma~\ref{Lemma:optimal:detector}), and then give the dual reformulation on the worst-case risk problem $\sup\{\Phi^*(\bP_0,\bP_1):~\bP_k\in\cP_k, k\in\bF\}$
\if\fullversion2
(see Lemma~\ref{Strong:Duality:Optimal:Detector} in Appendix~\ref{Appendix:proof}).
\fi
\if\fullversion1
We highlight that the reference \citep{gao18robust} has developed a similar result,
\fi
\if\fullversion2
We highlight that the reference \citep{gao18robust} has developed a similar result as in Lemma~\ref{Strong:Duality:Optimal:Detector},
\fi
in which the ambiguity sets are constructed using Wasserstein distance. 
However, their results cannot be directly applied because the LFDs of Wasserstein DRO are supported on finite number of points, so the dual problem is finite-dimensional and the duality of finite-dimensional convex programming holds.
In contrast, our dual problem is infinite-dimensional as the LFDs are absolutely continuous.
We leverage a non-trivial conic duality theorem in \citep[Theorem~2.165]{bonnans2013perturbation} to argue that the strong duality still holds.
Finally, we reformulate the inner supremum problem on the LHS of \eqref{Eq:relation:strong:duality} by applying the strong duality result of Sinkhorn DRO in Theorem~\ref{Theorem:Sinkhorn}, and then construct primal optimal solutions to show the duality gap between LHS and RHS in \eqref{Eq:relation:strong:duality} can be arbitrarily small.
\end{IEEEproof}

\subsection{Step 2: Finding Least Favorable Distributions}
Next, we discuss how to find LFDs by solving the following infinite-dimensional optimization problem
\begin{equation}\label{Eq:Psi:*}
\begin{aligned}
\sup_{\bP_k\in\cP, {k\in\bF}}&\quad \Phi^*(\bP_0,\bP_1)\\
\mbox{s.t.}&\quad \cW_{\Reg}(\hP_k, \bP_k)\le \rho_k, {k\in\bF}.
\end{aligned}
\end{equation}
The current formulation \eqref{Eq:Psi:*} is intractable because the decision variable is infinite-dimensional.
Moreover, it cannot be solved following standard tools from Sinkhorn DRO as the objective function $\Phi^*(\bP_0,\bP_1)$ is not linear with respect to $\bP_0$ and $\bP_1$.
To tackle this challenge, we first identify that this problem can be reformulated as a conic optimization problem with entropic constraints.

\begin{lemma}[Reformulation of \eqref{Eq:Psi:*}]\label{Lemma:reformulate:Phi*}
Under the setting of Proposition~\ref{Strong:Duality:inf:sup}, the problem \eqref{Eq:Psi:*} can be reformulated as
\begin{subequations}\label{Eq:Psi*:reformulated}
\begin{align}
\sup_{\substack{\ell_{i,k}\ge0,\\ i\in[n], k\in\bF}}&\quad 
\int \psi\left( 
\frac{\diff\bP_0}{\diff(\bP_0+\bP_1)}\right)\diff(\bP_0+\bP_1)\label{Eq:Psi*:reformulated:a}\\
\mbox{s.t.}&\quad \frac{\Reg}{n}\sum_{i=1}^n\int \ell_{i,k}(z)\log(\ell_{i,k}(z))\diff\bQ_{i,\Reg}^k(z)\le \orho_k,\label{Eq:Psi*:reformulated:b}\\
&\quad \int \ell_{i,k}(z)\diff\bQ_{i,\Reg}^k(z)=1, \label{Eq:Psi*:reformulated:c}\\
&\quad \diff\bP_k=\frac{1}{n}\sum_{i=1}^n\ell_{i,k}\diff\bQ_{i,\Reg}^k.\label{Eq:Psi*:reformulated:d}
\end{align}
\end{subequations}
\end{lemma}

To derive the reformulation \eqref{Eq:Psi*:reformulated}, we first apply the definition of Sinkhorn distance so that decision variables are the joint distributions between $\hP_k$ and $\bP_k$, denoted as $\gamma_k$, $k\in\bF$.
By the disintegration theorem, the joint distribution can be represented as $\gamma_k=\frac{1}{n}\sum_{i=1}^n\delta_{x_i^k}\otimes\gamma_{i,k}$,
where $\gamma_{i,k}$ stands for the conditional distribution of $\gamma_k$ given the first marginal of $\gamma_k$ equals $x_i^k$.
Define the importance ratio function $\ell_{i,k}:~\Omega\to\bR_+$ as $\ell_{i,k}(z)=\diff\gamma_{i,k}(z)/\diff\bQ_{i,\Reg}^k(z)$. Substituting the expressions of $\orho_k$ and $\bQ_{i,\Reg}^k$ implies the desired formulation.

\begin{remark}[Interpretation of Sinkhorn Detector]
Constraint of the problem \eqref{Eq:Psi:*} can also be reformulated as
\begin{align*}
\frac{\Reg}{n}\sum_{i=1}^nD_{\text{KL}}(\gamma_{i,k}\|\bQ_{i,\Reg}^k)&\le \orho_k,\quad
\bP_k = \frac{1}{n}\sum_{i=1}^n\gamma_{i,k},
\end{align*}
where $\gamma_{i,k}$ is the conditional transport mapping provided that the first marginal equals to $x_i^k$.
In other words, Sinkhorn DRO formulation \eqref{Eq:Psi:*} can be understood as a {generalized KL-divergence constrained} problem.
When $\orho_k=0$ for $k\in\bF$, the constraint set only contains one feasible solution
$
\overline{\bP}_k = \frac{1}{n}\sum_{i=1}^n\bQ_{i,\Reg}^k,
$
which can be viewed as the non-parametric smooth density estimation constructed from samples $\{x_i^{k}\}_i$.
Consequently the optimal detector is the one based on estimated densities $\overline{\bP}_0$ and $\overline{\bP}_1$.
\end{remark}

The support of decision variables $\ell_{i,k}$ is the same as $\text{supp}(\bQ_{i,\Reg}^k)$, making the reformulated problem \eqref{Eq:Psi*:reformulated} still infinite-dimensional and therefore intractable.
We solve its sample estimate problem instead, leveraging the Monte-Carlo approximation.
For each $i$ and $k$, we sample $m$ points from $\bQ_{i,\Reg}^k$ and denote the corresponding empirical distribution as $\hat{\bQ}_{i,\Reg}^k$. 
If directly replacing the kernel distribution $\bQ_{i,\Reg}^k$ with its empirical counterpart for the formulation in \eqref{Eq:Psi*:reformulated}, the LFDs $\bP_0$ and $\bP_1$ will have non-overlapping supports, and consequently the optimal detector is not well-defined.
We leverage the idea of importance sampling to derive the Monte-Carlo approximated problem.
Define the probability measure $\hat{\mathbb{G}}_{i,\Reg}$ as $\hat{\mathbb{G}}_{i,\Reg} = (\hat{\bQ}_{i,\Reg}^0 + \hat{\bQ}_{i,\Reg}^1)/2$, and let $r_{i,\Reg}^k:~\Omega\to\bR_+$ be the weighted importance ratio function between the kernel distributions:
\[
r_{i,\Reg}^k(z):=\frac{2\diff\bQ_{i,\Reg}^k}{\diff(\bQ_{i,\Reg}^0 + \bQ_{i,\Reg}^1)}(z).
\]
As a consequence, the problem \eqref{Eq:Psi*:reformulated} can be approximated as a finite-dimensional optimization problem:
\begin{subequations}\label{Eq:Psi*:approximated}
\begin{align}
\sup_{\substack{\ell_{i,k}\in\bR_+^{2m},\\ i\in[n], k\in\bF}}&\quad 
\int \psi\left( 
\frac{\diff\bP_0}{\diff(\bP_0+\bP_1)}\right)\diff(\bP_0+\bP_1)\\
\mbox{s.t.}&\quad \frac{\Reg}{n}\sum_{i=1}^n\int \ell_{i,k}\log(\ell_{i,k})r_{i,\Reg}^k\diff\hat{\mathbb{G}}_{i,\Reg}\le \orho_k,\label{Eq:Psi*:approximated:b}\\
&\quad \int \ell_{i,k}r_{i,\Reg}^k\diff\hat{\mathbb{G}}_{i,\Reg}=\int r_{i,\Reg}^k\diff\hat{\mathbb{G}}_{i,\Reg},\label{Eq:Psi*:approximated:c}\\
&\quad \diff\bP_k=\frac{\frac{1}{n}\sum_{i=1}^n\ell_{i,k}r_{i,\Reg}^k\diff \hat{\mathbb{G}}_{i,\Reg}}{ 
\frac{1}{n}\sum_{i=1}^n\int r_{i,\Reg}^k\diff\hat{\mathbb{G}}_{i,\Reg}
}.\label{Eq:Psi*:approximated:d}
\end{align}
\end{subequations}
The approximated problem \eqref{Eq:Psi*:approximated} always contains a feasible solution $\ell_{i,k}=1, i\in[n],k\in\bF$.
In addition, constraints \eqref{Eq:Psi*:approximated:b}-\eqref{Eq:Psi*:approximated:d} are consistent estimates of the constraints \eqref{Eq:Psi*:reformulated:b}-\eqref{Eq:Psi*:reformulated:d}, respectively.
It is an open question whether the optimal value of the approximated problem~\eqref{Eq:Psi*:approximated} is a consistent estimate of the optimal value in \eqref{Eq:Psi:*}.
The technical difficulty is due to the infinite problem size of \eqref{Eq:Psi:*} so that discussions on properties of sample approximation estimators in \citep[Section~5.1]{shapiro2021lectures} do not apply.
We hope to address this issue in future works.

Since the importance ratio $\ell_{i,k}$ is supported on $\text{supp}(\hat{\mathbb{G}}_{i,\Reg})$, which consists of $2m$ points, the LFDs $\bP_0^*$ and $\bP_1^*$ from~\eqref{Eq:Psi*:approximated} will have the common support, consisting of $2mn$ points.
The approximated problem can be efficiently solved using common off-the-shelf software such as CVX~\citep{cvx,gb08}.
\if\fullversion2
In addition, we provide visualization of LFDs and impact of regularization parameters using a toy example in Appendix~\ref{Appendix:visual}.
\fi
\begin{remark}[$k$-NN Detector]\label{Remark:kNN:detector}
When making inference on any target sample $\omega$ that is beyond the support of $\bP_0^*$ and $\bP_1^*$, the approximated detector is defined using a weighted k-NN classifier:
\[
\tilde{T}(\omega) = \frac{1}{K}\sum_{s=1}^{K}~q_sT^*(x_s^*),
\]
where $x_1^*,\ldots,x_K^*$ are the $K$ nearest neighbors of $\omega$ and supported on $\bP^*_k$, $k\in\bF$, and $q_s$ is inversely proportional to $\|x_s^*-\omega\|$. 
We take $K=5$ during numerical simulations.
\end{remark}

\begin{remark}[Complexity of \eqref{Eq:Psi*:approximated}]
The complexity of solving \eqref{Eq:Psi*:approximated} is independent of the data dimension $d$, as we only require the importance ratio functions evaluated on samples from $\hat{\mathbb{G}}_{i,\Reg}, i\in[n]$ as inputs to the convex program.
Moreover, as the constraint set is a ball of weighted $\ell_1$-norm, from convex optimization theory~\citep{nemirovski2001lectures} we know that when the objective is Lipschitz in $\ell_1$-norm, the computational complexity is of $O(\log(mn))$, which is nearly sample size independent.
This is true for all except the first case in Table~\ref{tab:generating:function}.
\end{remark}

\section{Applications}\label{Sec:application}
In this section, we apply our proposed method in three applications: composite hypothesis testing, digits classification, and change-point detection.
We take the cost function $c(x,y)=\frac{1}{2}\|x-y\|_2^2$, and the reference measure $\nu$ for Sinkhorn distance is chosen to be the Lebesgue measure.
For benchmark comparison, we also report the performance for other tests such as Wasserstein robust test~\citep{xie2021robust}, MMD robust test~\citep{Zhongchang21}, and neural network classification logit test~\citep{cheng2020classification}.
Hyper-parameters such as the radii of uncertainty sets and the entropic regularization parameter are selected using cross validation.
\if\fullversion1
Other experiment details are omitted in \citep[Appendix~\ref{Appendix:experiment}]{jie22isit_full}.
\fi
\if\fullversion2
Other experiment details are omitted in Appendix~\ref{Appendix:experiment}.
\fi
\if\submitversion1
\begin{figure}[t!]
    \centering
    \includegraphics[height=0.165\textwidth]{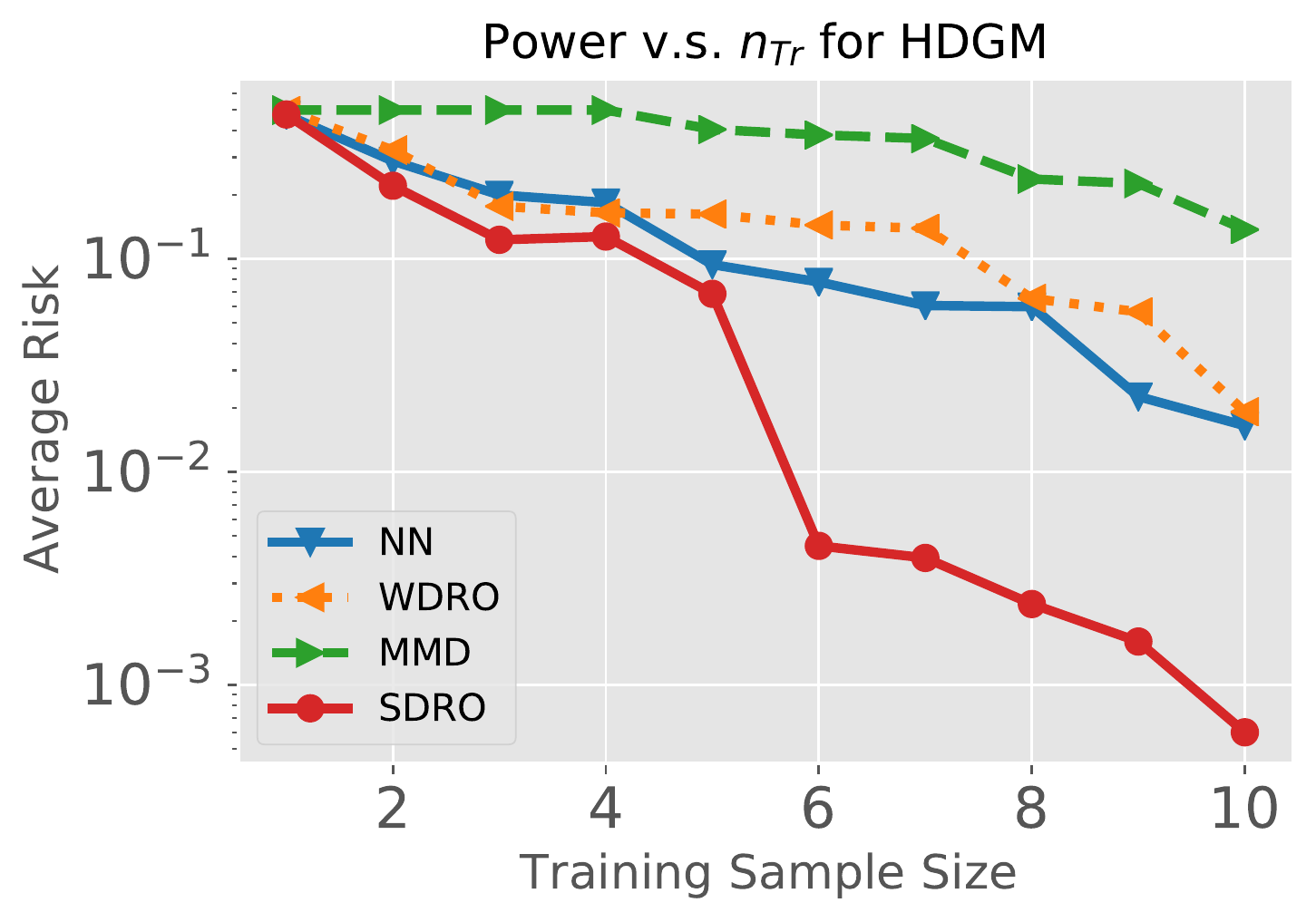}
    \includegraphics[height=0.165\textwidth]{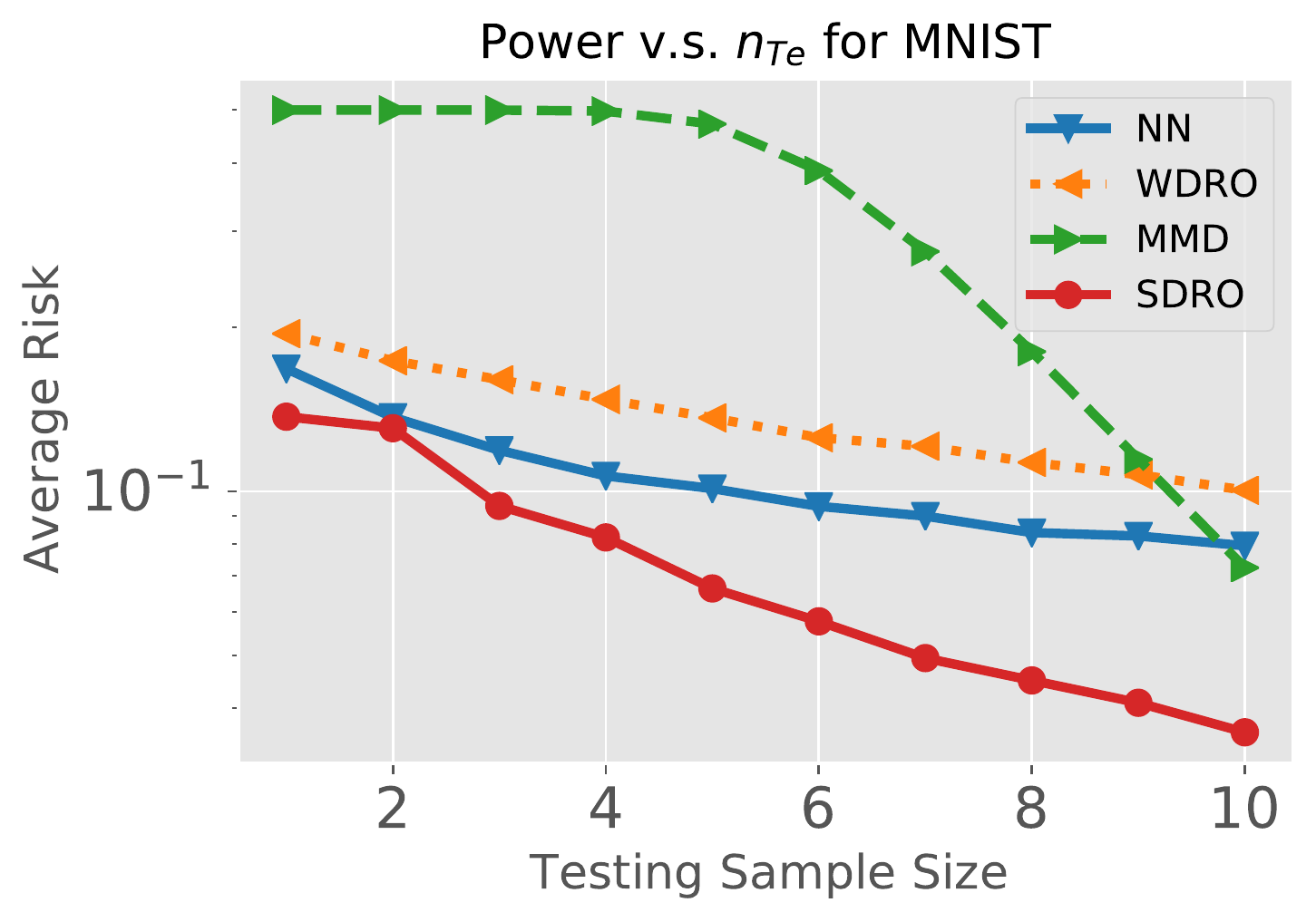}
    \caption{Testing risk for HDGM Data (left) and MNIST Data (right).}
    \label{fig:risk:HDGM}
\end{figure}
\fi
\if\submitversion2
\begin{figure}[t!]
    \centering
    \includegraphics[height=0.3\textwidth]{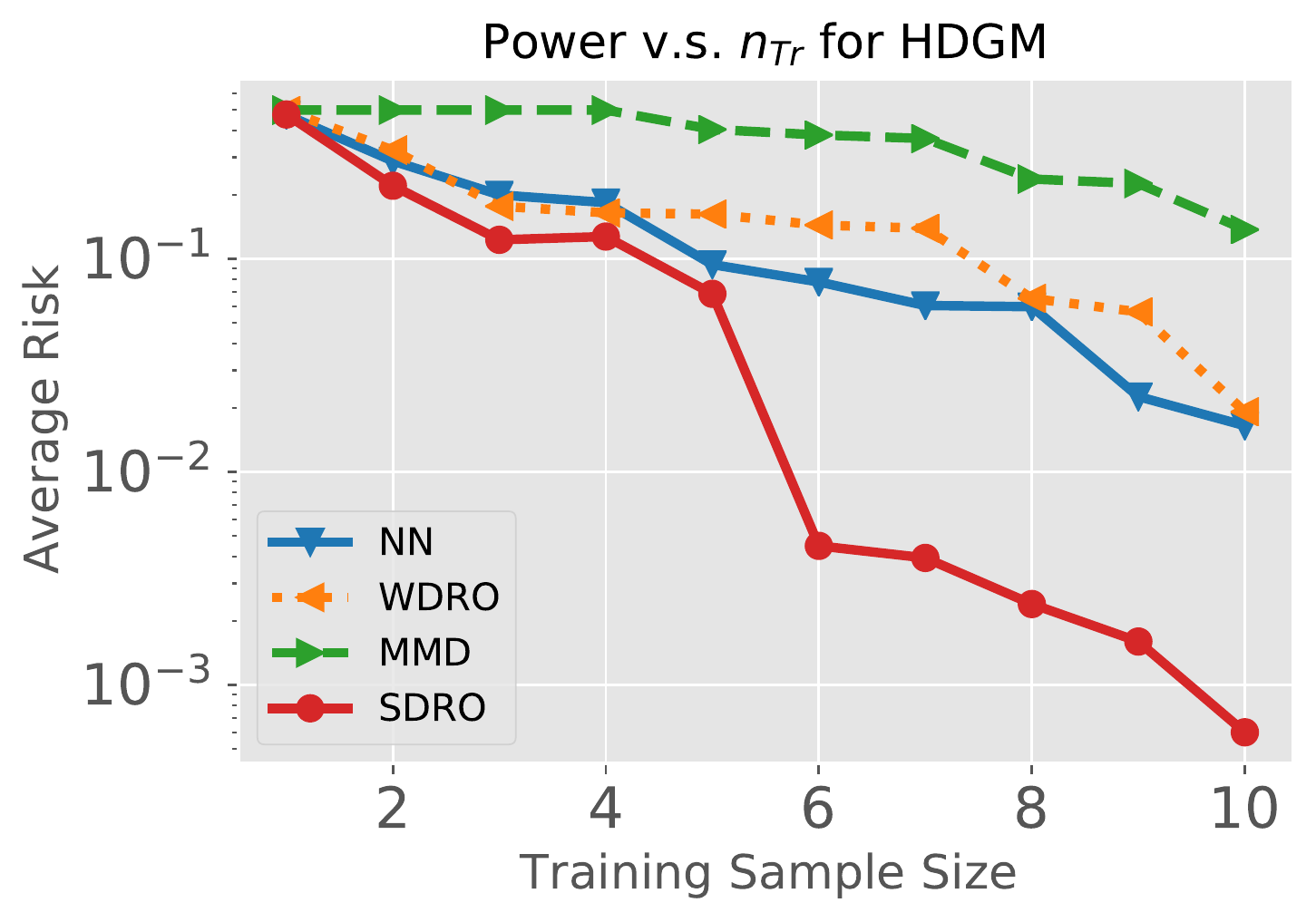}
    \includegraphics[height=0.3\textwidth]{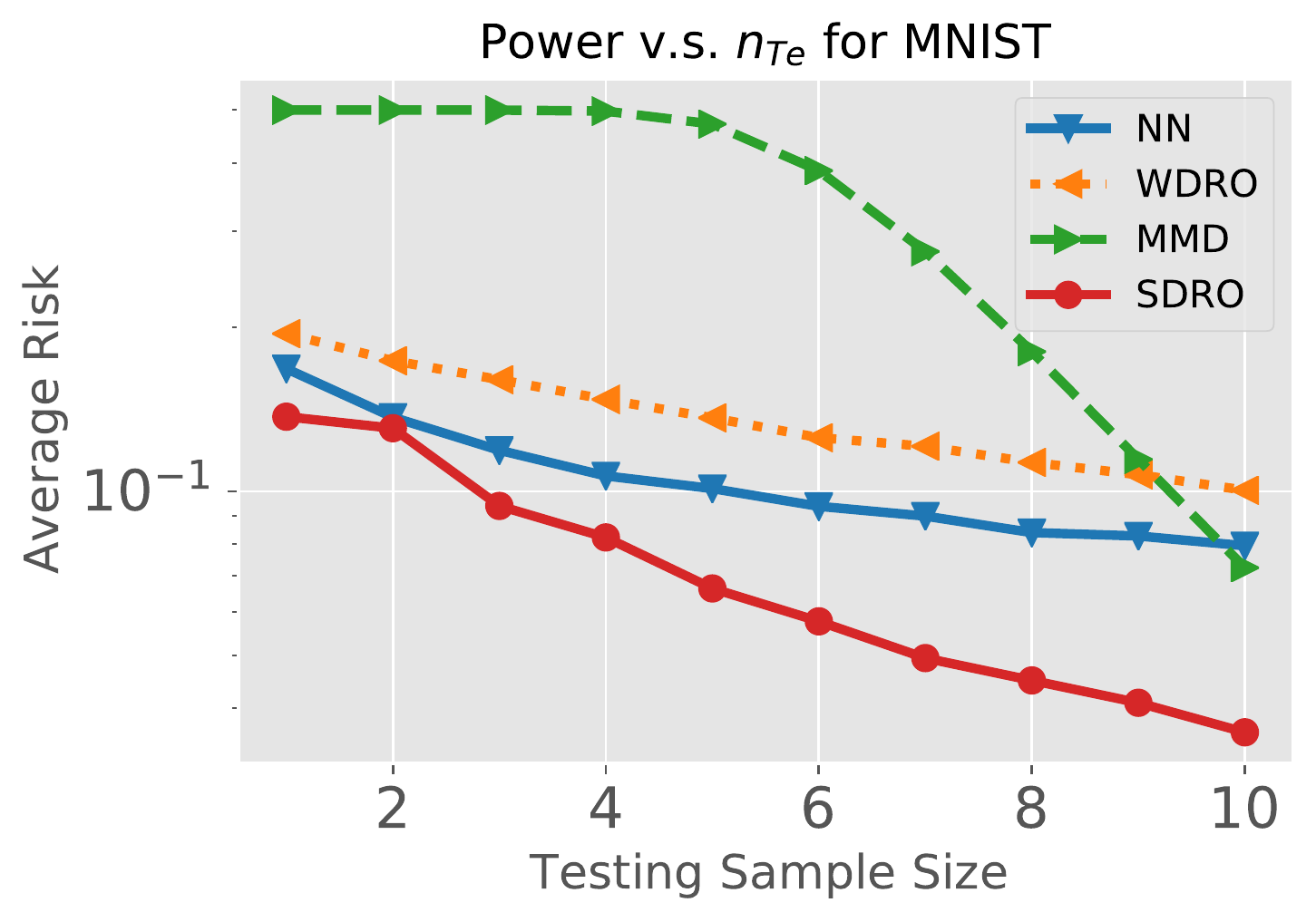}
    \caption{Testing risk for HDGM Data (left) and MNIST Data (right).}
    \label{fig:risk:HDGM}
\end{figure}
\fi

\subsection{Composite Hypothesis Testing}\label{Sub:composite:hypothesis}
Assume samples from two hypotheses are generated from high dimensional Gaussian mixture models~(HDGM) following distributions $\sum_{i=1}^2\frac{1}{2}\cN((-1)^ie,I_D)$ and $\sum_{i=1}^2\frac{1}{2}\cN((-1)^if,I_D)$, respectively, where $D=100$, $e$ is the unit vector in $\bR^D$, and $f$ is a vector with the first half entries being $1$ and the remaining half being $-1$.
We find the optimal detectors based on $n\in[10]$ training samples from each distribution.
Then we test its averaged mis-classification rates based on $1000$ new testing samples from each distribution.
Then this experiment is repeated for $10$ independent trials.
Experiment results for this part are reported in Fig.~\ref{fig:risk:HDGM}, from which we can see that our proposed method performs the best over others, suggesting that it is useful for small-sample scenarios.

\subsection{MNIST Digits Classification}
Next, we examine the performance in the task of digits classification.
We randomly select five images from the MNIST dataset~\citep{lecun10} for digits $1$ and $2$ as training samples.
Then we divide test images from the same class into batches, each consisting of $n_{\text{Te}}\in[10]$ samples.
We compute the mis-classification rates for $1000$ randomly selected batches, and repeat the experiment for $10$ independent trials.
Experiment results are reported in Fig.~\ref{fig:risk:HDGM}, from which we can see that the risk of our proposed method decays quickly into zero as the testing batch size increases, and it significantly outperforms the others.

\begin{table}[t!]
  \centering
  \footnotesize
  \caption{
  Detection power for the task of change-point detection in four synthetic datasets.
  For each instance the experiment is repeated for $100$ independent trials.
  Thresholds for all methods are calibrated so that the significance level is $\alpha=0.05$.
  }
  \vspace{1mm}
  \rowcolors{2}{white}{gray!30}
    \begin{tabular}{p{1cm}p{1cm}p{1cm}p{1cm}p{1cm}}
    \toprule
    & Case 1 & Case 2 & Case 3 & Case 4\\
 \midrule
 NN & 0.12       & 0.40       & 0.37 & {\bf 0.58}\\
WDRO& 0.66       & 0.75       & 0.42       & 0.45\\
SDRO& {\bf 0.69} & {\bf 0.82} & {\bf 0.53}       &0.56\\
    \bottomrule
    \end{tabular}%
  \label{tab:MSRC:full}%
  \vspace{-1em}
\end{table}%

\subsection{Offline Change-point Detection}
Finally, we investigate the performance for the offline change-point detection.
Suppose a series of samples are given with time horizon $T=200$ and we set the change point $K=100$.
The goal is to detect the change-point based on given samples.

The detection procedure is as follows. Take a sliding window size $\omega=20$.
For any candidate change time $t$, we treat samples from $[t-\omega,t-1]$ and $[t+\omega,t]$ as two groups of observations and solve for the LFDs $\bP_0^*$ and $\bP_1^*$, based on which we calculate the detection statistics as $D_t=-\tilde{T}(\omega_t)$.
We compute the CUSUM-type~\citep{Liyanchange_21} recursive detection statistic $S_t=\max\{0, S_{t-1} + D_t\}$.
A change is detected if $S_t$ exceeds a pre-specified threshold. 
Thresholds for all methods are calibrated so that the false alarm rate is controlled within $\alpha=0.05$.
\if\fullversion2
We consider four cases of distribution changes using synthetic dataset, and the details are deferred in Appendix~\ref{Appendix:experiment}.
\fi
\if\fullversion1
We consider four cases of distribution changes using synthetic dataset, and the details are deferred in \citep[Appendix~\ref{Appendix:experiment}]{jie22isit_full}.
\fi

Table~\ref{tab:MSRC:full} reports the testing power, i.e., the probability of successfully detecting a change when the change exists, for various methods averaged for $100$ independent trials.
It shows that Sinkhorn robust test can capture the difference between pre- and post-change distributions well except that for the last case, NN slightly outperforms the Sinkhorn test.

\section{Conclusion}\label{Sec:conclusion}
We developed a data-driven approach for the problem of robust hypothesis testing in sample-sample scenario.
In particular, we proposed a distributionally robust optimization formulation that optimizes the worst-case risk over all distributions within ambiguity sets using Sinkhorn distance.
Generalizing this approach into other settings such as type-I error constrained tests or multiple hypothesis tests could be of research interest.

\if\submitversion1
\twocolumn
\balance
\clearpage
\bibliographystyle{IEEEtran}
\bibliography{shortbib.bib}
\fi

\if\submitversion2
\bibliographystyle{IEEEtran}
\bibliography{shortbib.bib}
\fi

\onecolumn
\appendices

\section{Proofs of Technical Results}\label{Appendix:proof}

\begin{IEEEproof}[Proof of Lemma~\ref{Lemma:reformulate:Phi*}]
Leveraging Definition~\ref{Def:Sinkhorn}, we reformulate the problem \eqref{Eq:Psi:*} as
\begin{subequations}
\begin{align}
\sup_{\gamma_k\in\cP(\Omega\times\Omega), \bP_k\in\cP(\Omega), k\in\bF}~&\quad \int \psi\left( 
\frac{\diff\bP_0}{\diff(\bP_0+\bP_1)}
\right)\diff(\bP_0+\bP_1)
\\
\mbox{s.t.}&\quad
\mathbb{E}_{\gamma_k}\left[ 
c(x,z) + \Reg\log\left( 
\frac{\diff\gamma(x,z)}{\diff\hP_k(x)\diff\nu(z)}
\right)
\right]\le \rho_k,\quad k\in\bF\label{Eq:constranint:b:12}
\\&\quad \Proj_{1\#}\gamma_k=\hP_k, \Proj_{2\#}\gamma_k=\bP_k, \quad k\in\bF,\label{Eq:constranint:c:12}
\end{align}
\end{subequations}
where $\Proj_{1\#}\gamma$ and $\Proj_{2\#}\gamma$ are the first and the second marginal distributions of $\gamma$, respectively.
By the disintegration theorem~\citep{chang1997conditioning}, we reformulate the joint distribution 
\begin{equation}\label{Expression:joint:gamma_k}
\gamma_k=\frac{1}{n}\sum_{i=1}^n\delta_{x_i^k}\otimes\gamma_{i,k}, 
\end{equation}
where $\gamma_{i,k}$ is the conditional distribution of $\gamma_k$ given the first marginal of $\gamma_k$ equals $x_i^k$.
As a consequence, the constraint \eqref{Eq:constranint:b:12} becomes
\[
\frac{1}{n}\sum_{i=1}^n
\mathbb{E}_{\gamma_{i,k}}\left[ 
c(x_i^k,z) + \Reg\log\left( 
\frac{\diff\gamma_{i,k}(z)}{\diff\nu(z)}
\right)
\right]\le \rho_k,\quad k\in\bF.
\]
Substituting the expression of $\orho_k$ and $\bQ_{i,\Reg}^{k}$ defined in \eqref{Eq:orho:k} and \eqref{Eq:Q:i:Reg:k} into the equation above, the constraint \eqref{Eq:constranint:b:12} can be reformulated as
\[
\frac{1}{n}\sum_{i=1}^n\mathbb{E}_{\gamma_{i,k}}\left[ 
\log\left(
\frac{\diff\gamma_{i,k}(z)}{\diff\bQ_{i,\Reg}^{k}(z)}
\right)
\right]\le \orho_k,\quad k\in\bF.
\]
Now define the importance ratio function $\ell_{i,k}:~\Omega\to\bR_+$ as
\[
\ell_{i,k}(z) = \frac{\diff\gamma_{i,k}}{\diff\bQ_{i,\Reg}^{k}}(z),\quad \forall z\in\Omega,
\]
then the relation \eqref{Eq:constranint:b:12} is equivalent to 
\begin{equation}
\frac{1}{n}\sum_{i=1}^n \int \ell_{i,k}(z)\log(\ell_{i,k}(z))\diff\bQ_{i,\Reg}^k(z)\le \orho_k,\quad k\in\bF.
\label{Eq:reformulate:constranint:b:12}
\end{equation}
From the expression of $\gamma_k$ in \eqref{Expression:joint:gamma_k}, we realize that
\begin{equation}\label{Eq:expression:bPk}
\diff\bP_k = \frac{1}{n}\sum_{i=1}^n\diff\gamma_{i,k} = \frac{1}{n}\sum_{i=1}^n
\ell_{i,k}\diff\bQ_{i,\Reg}^k,\quad k\in\bF.
\end{equation}
Combining expressions \eqref{Eq:reformulate:constranint:b:12} and \eqref{Eq:expression:bPk}, we derive the desired reformulation.

\end{IEEEproof}

\begin{lemma}[Strong Duality for Optimal Detector]\label{Strong:Duality:Optimal:Detector}
Under the setting of Proposition~\ref{Strong:Duality:inf:sup}, suppose the radii defined in \eqref{Eq:orho:k} satisfy $\orho_k>0, k\in\bF$, then the strong duality holds for the optimal detector problem: 
\begin{align*}
&\underbrace{\sup_{\substack{\bP_k\in\cP_k,\\ {k\in\bF}}}~\int [\psi\circ r(\omega)]\diff(\bP_0+\bP_1)(\omega)}_{\Psi^*}\\
=&\inf_{\substack{\lambda_k\ge0,\\ {k\in\bF}}}\sup_{
\substack{\ell_{i,k}\ge0,\\
\int \ell_{i,k}(z)\diff\bQ_{i,\Reg}^k(z)=1,\\
i\in[n],
k\in\bF
}}
\mathcal{L}\left(\{\lambda_k\}_{k\in\bF}, \{\ell_{i,k}\}_{i\in[n],k\in\bF}\right),
\end{align*}
where the Lagrangian function $\mathcal{L}$ is defined as
\begin{equation}\label{Eq:lagrangian:optimal:detector}
\begin{aligned}
&\mathcal{L}(\{\lambda_k\}_k, \{\ell_{i,k}\}_{i,k})=
\sum_{k=0}^1\lambda_k\left[\orho_k
-\frac{\Reg}{n}\sum_{i=1}^n\int \ell_{i,0}(z)\log(\ell_{i,0}(z))\diff\bQ_{i,\Reg}^k(z)
\right]\\
&\qquad\qquad\qquad\qquad
+
\int \psi\left(
\frac{\frac{1}{n}\sum_{i=1}^n\ell_{i,0}\diff\bQ_{i,\Reg}^0}{\frac{1}{n}\sum_{i=1}^n(\ell_{i,0}\diff\bQ_{i,\Reg}^0 + \ell_{i,1}\diff\bQ_{i,\Reg}^1)}(z)
\right)\frac{1}{n}\sum_{i=1}^n(\ell_{i,0}\diff\bQ_{i,\Reg}^0 + \ell_{i,1}\diff\bQ_{i,\Reg}^1)(z).
\end{aligned}
\end{equation}
Moreover, the saddle point solution for the minimax problem above is guaranteed to exist. 
\end{lemma}
\begin{IEEEproof}[Proof of Lemma~\ref{Strong:Duality:Optimal:Detector}]
From Lemma~\ref{Lemma:reformulate:Phi*}, we reformulate the optimal value $\Phi^*$ as
\begin{align*}
\sup_{
\substack{\ell_{i,k}\ge0,\\
\int \ell_{i,k}(z)\diff\bQ_{i,\Reg}^{k}(z)=1,\\
i\in[n],
k\in\bF
}}&\quad 
\int \psi\left(
\frac{\diff \bP_0}{\diff(\bP_0+\bP_1)}
\right)\diff(\bP_0+\bP_1)\\
\mbox{s.t.}&\quad \frac{\Reg}{n}\sum_{i=1}^n\int \ell_{i,k}(z)\log(\ell_{i,k}(z))\diff\bQ_{i,\Reg}^k(z)\le \bar{\rho}\\
&\quad \bP_k=\frac{1}{n}\sum_{i=1}^n\ell_{i,k}\bQ_{i,\Reg}^k,\quad k\in\bF.
\end{align*}
We can see that the Slater's condition holds by taking $\ell_{i,k}=1_{\Omega}$, and the functional 
\[
\{\ell_{i,k}\}_{i\in[n]}\to \frac{\Reg}{n}\sum_{i=1}^n\int \ell_{i,k}(z)\log(\ell_{i,k}(z))\diff\bQ_{i,\Reg}^k(z)
\]
is lower semi-continuous for $k\in\bF$.
Applying \citep[Theorem~2.165]{bonnans2013perturbation} implies that $\Psi^*$ has the strong dual reformulation, and the saddle point solution corresponding to the Lagrangian function \eqref{Eq:lagrangian:optimal:detector} is guaranteed to exist.

\end{IEEEproof}

\begin{IEEEproof}[Proof of Proposition~\ref{Strong:Duality:inf:sup}]
When $\orho_k=0$ for $k\in\bF$, uncertainty sets $\cP_0$ and $\cP_1$ only contain one singleton, making the relation \eqref{Eq:relation:strong:duality} trivially holds.
In the following we focus on the case where $\orho_k>0, k\in\bF$, while cases $\orho_0=0,\orho_1>0$ and $\orho_0>0,\orho_1=0$ can be handled in a similar manner.
By the minimax inequality, exchanging inf and sup in \eqref{Eq:minimax:test} yields 
\[
\inf_{T:~\Omega\to\bR}~\sup_{\substack{\bP_k\in\cP_k, \\{k\in\bF}}}~\Phi(T;\bP_0,\bP_1)
\ge
\sup_{\substack{\bP_k\in\cP_k, \\{k\in\bF}}}\inf_{T:~\Omega\to\bR}~\Phi(T;\bP_0,\bP_1).
\]
It suffices to show that 
\[
\inf_{T:~\Omega\to\bR}~\sup_{\substack{\bP_k\in\cP_k, \\{k\in\bF}}}~\Phi(T;\bP_0,\bP_1)\le \Psi^*,
\]
where the optimal value $\Psi^*$ is defined in Lemma~\ref{Strong:Duality:Optimal:Detector}.
Take
\[
c^{\gamma,\Reg}_k(x,z)=c(x,z) + \Reg\log\left( 
\frac{\diff\gamma(x,z)}{\diff \hP_k(x)\diff\nu(z)}
\right).
\]
Leveraging the strong duality result for Sinkhorn DRO in Theorem~\ref{Theorem:Sinkhorn}, for any fixed $T$ it holds that
\begin{align*}
\Phi^*(T)&\triangleq\sup_{\bP_k\in\cP_k, {k\in\bF}}~\Phi(T;\bP_0,\bP_1)\\
&=\inf_{\substack{\lambda_k\ge0, \text{Proj}_{1\#\gamma_k}=\hP_k\\{k\in\bF}}}~
D(T;\{\lambda_k\}, \{\gamma_k\}),
\end{align*}
where
\begin{align*}
&D(T;\{\lambda_k\}, \{\gamma_k\})=\lambda_0\rho_0+\lambda_1\rho_1 \\
&\qquad\qquad +\int\left[\ell(-T(z_0)) - \lambda_0c^{\gamma_0,\Reg}_0(x_0,z_0)\right]\diff\gamma_0(x_0,z_0)\\
&\qquad\qquad +\int\left[\ell(T(z_1)) - \lambda_1c^{\gamma_1,\Reg}_1(x_1,z_1)\right]\diff\gamma_1(x_1,z_1).
\end{align*}
We now construct the approximate primal optimal solution.
Denote by $(\{\lambda_k^*\}_k,\{\ell^*_{i,k}\}_{i,k})$ the saddle point optimal solution for the optimal value $\Psi^*$.
We construct the transport mapping $\gamma_k^*$ for ${k\in\bF}$ such that 
\[
\diff\gamma_k^*(x,z) = \frac{1}{n}\sum_{i=1}^n\ell^*_{i,k}(z)\diff\delta_{x_i^k}(x)\diff\bQ_{i,\Reg}^k(z).
\]
Because of the sub-optimality of $(\{\lambda_k^*\}_k,\{\ell^*_{i,k}\}_{i,k})$, for any detector $T$ it holds that
\begin{equation}
\Phi^*(T)\le
D(T;\{\lambda_k^*\}, \{\gamma_k^*\})
\triangleq\mathcal{L}_{T}(\{\lambda_k^*\}_k, \{\ell_{i,k}^*\}_{i,k}),
\label{Eq:suboptimality:lambda:ell}
\end{equation}
where by substitution, the term $\mathcal{L}_{T}(\{\lambda_k^*\}_k, \{\ell_{i,k}^*\}_{i,k})$ is defined as
\begin{equation}\label{Eq:lagrangian:near:optimal:detector}
    \begin{aligned}
        &\mathcal{L}_{T}(\{\lambda_k\}_k, \{\ell_{i,k}\}_{i,k})=\sum_{k=0}^1\lambda_k\left[\orho_k
-\frac{\Reg}{n}\sum_{i=1}^n\int \ell_{i,0}(z)\log(\ell_{i,0}(z))\diff\bQ_{i,\Reg}^k(z)
\right]\\
&\qquad\qquad
+\int \psi_{T(z)}\left(
\frac{\frac{1}{n}\sum_{i=1}^n\ell_{i,0}\diff\bQ_{i,\Reg}^0}{\frac{1}{n}\sum_{i=1}^n(\ell_{i,0}\diff\bQ_{i,\Reg}^0 + \ell_{i,1}\diff\bQ_{i,\Reg}^1)}(z)
\right)\frac{1}{n}\sum_{i=1}^n(\ell_{i,0}\diff\bQ_{i,\Reg}^0 + \ell_{i,1}\diff\bQ_{i,\Reg}^1)(z),
    \end{aligned}
\end{equation}
and $\psi_t(r)\triangleq (1-r)\ell(t) + r\ell(-t), r\in [0,1], t\in\bR$.
On the other hand, from Lemma~\ref{Strong:Duality:Optimal:Detector} we can see
\begin{align*}
\Psi^*=&\mathcal{L}(\{\lambda_k^*\}_k, \{\ell_{i,k}^*\}_{i,k}).
\end{align*}
Comparing the expression of $\mathcal{L}$ in \eqref{Eq:lagrangian:optimal:detector} and $\mathcal{L}_{T}$ in \eqref{Eq:lagrangian:near:optimal:detector}, we have that for any $\delta>0$, there exists a detector $T_{\delta}$ so that 
\[
\Phi^*(T_{\delta})\le 
\mathcal{L}_{T_{\delta}}(\{\lambda_k^*\}_k, \{\ell_{i,k}^*\}_{i,k})
\le
\mathcal{L}(\{\lambda_k^*\}_k, \{\ell_{i,k}^*\}_{i,k})+\delta
=
\Psi^* + \delta.
\]
Taking $\delta\to0$ completes the proof.
\end{IEEEproof}

\section{Experimental Details and Additional Results}\label{Appendix:experiment}

\subsection{Procedure of Cross Validation}
To select hyper-parameters for all methods, we randomly partition the given samples into a training set with $50\%$ data and a validation set with the remaining data.
We obtain detectors using the training set across different choices of hyper-parameters and choose the one with the smallest mis-classification risk using the validation set.
The exception is that for robust MMD test, we train detectors using all the data and post select the hyper-parameter with the best performance.

\subsection{Detailed Procedure of Benchmark Methods}
When using neural network-based testing, we parameterize the detector function and optimize the corresponding objective function:
\[
\min_{\bm\theta}~\mathbb{E}_{\hP_0}~[\ell\circ(-T_{\bm\theta})(\omega)]
+
+ \mathbb{E}_{\hP_1}[\ell\circ T_{\bm\theta}(\omega)],
\]
where $\ell(t)=\log(1+\exp(t))/\log 2$, and the detector $T_{\theta}$ is the output of the two-layer neural network:
\begin{equation*}
T_{\bm\theta}(x) = \frac{1}{N}\sum_{i=1}^N\sigma_*(x; \bm\theta[i]),\quad 
\bm\theta=(\bm\theta[i])_{i=1}^N.
\end{equation*}
In particular, we specify the activation function $\sigma_*(x; \theta)=a\sigma(w\cdot x+ b), \theta=(a,w,b)$, with $\sigma(\cdot)$ being the sigmoid operator.
We train the neural network using stochastic gradient descent with the number of neurons $N=200$ and the number of iterations $T=80$.

When using Wasserstein DRO-based testing, we obtain the detector function valued on training samples according to the formulation in \citep[Theorem 3]{gao18robust} and then obtain the detector valued on the testing sample $\omega$ according to Remark~\ref{Remark:kNN:detector}.
When using MMD DRO-based testing, we choose the Gaussian kernel with the bandwidth to be tuned, and obtain the detector function according to the formulation in \citep[Eq.~(14)]{Zhongchang21}.

\subsection{Datasets for Offline Change-Point Detection}

The experiment of offline change-point detection includes the following cases:
\begin{enumerate}
    \item 
(Discrete distributions).
The support size is $n=10$. Distribution shifts from the uniform distribution $\bP=\bm1/10$ to $\bQ=[1/60,2/60,3/60,1/5,1/5,1/5,1/5,3/60,2/60,1/60]$, a non-uniform distribution.
    \item
(Gaussian to Gaussian mixture).
Distribution shifts from $\cN(0,I_{20})$ to Gaussian mixture $0.8\cN(0,I_{20})+0.2\cN(0,0.1I_{20})$.
    \item
(Gaussian mean and covariance shift).
Distribution shifts from $\cN(0,I_2)$ to $\cN(\mu,\Sigma)$ with $\mu=(1,0)^T$ and $\Sigma=[0.5,0.1;0.1,0.5]$.
    \item
(Gaussian to Laplace).
Distribution shifts from $\cN(0,1)$ to Laplace distribution with mean $1$ and standard deviation $0.5$.
\end{enumerate}

\subsection{Visualization of Least Favorable Distributions}\label{Appendix:visual}

\begin{figure}[H]
    \centering
    \includegraphics[height=0.3\textwidth]{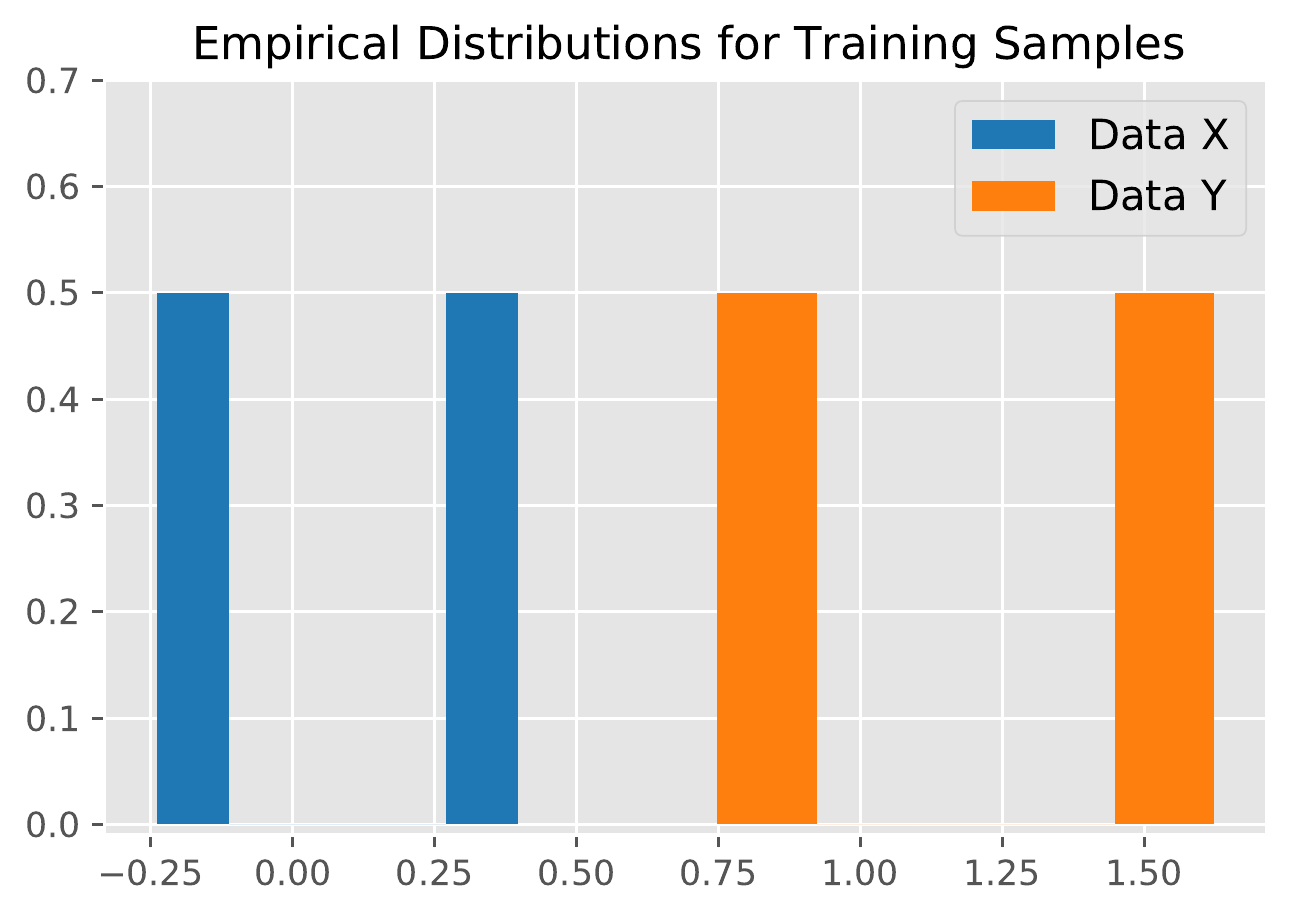}
    \caption{Plot for empirical distributions from $\{x_1^0,x_2^0\}$ and $\{x_1^1, x_2^1\}$.}
    \label{fig:histogram:toy}
\end{figure}

We provide visualization of LFDs using a toy example.
We generate $n=2$ samples from distributions $\bP_0=\cN(-0.1,1)$ and $\bP_1=\cN(0.1,1)$, respectively.
Samples for hypothesis $H_0$ are $x_1^0=0.39, x_2^0=-0.23$, 
and for hypothesis $H_1$ are $x_1^1=0.74, x_2^1=1.62$.
The plot for corresponding empirical distributions is presented in Fig.~\ref{fig:histogram:toy}. 
Specifically, we visualize LFDs based on those samples using WDRO test and SDRO test in Fig.~\ref{fig:LFD}.
The radii for Wasserstein ambiguity sets are set to be $\theta_0=\theta_1=0.19$, and that for Sinkhorn ambiguity sets are set to be $\orho_0=\orho_1=0.03$.
We take the number of Monte-Carlo approximations from $\{\bQ_{i,\Reg}^k\}$ to be $m=1000$, and we try different regularization parameters $\Reg\in\{0.01,0.1,1\}$
when using Sinkhorn distance.
From the plot we can see that the supports of LFDs are limited to training samples when using WDRO test.
In contrast, supports of LFDs for SDRO test are more flexible and usually beyond training samples. 
When using a relatively small regularization parameter, e.g., $\Reg=0.01$, the corresponding LFDs tend to support near training samples.
When using a large regularization parameter, e.g., $\Reg=1$, the LFDs tend to spread over the whole sample space $\bR$.
\if\submitversion1
\begin{figure}[H]
    \centering
    \includegraphics[width=0.24\textwidth]{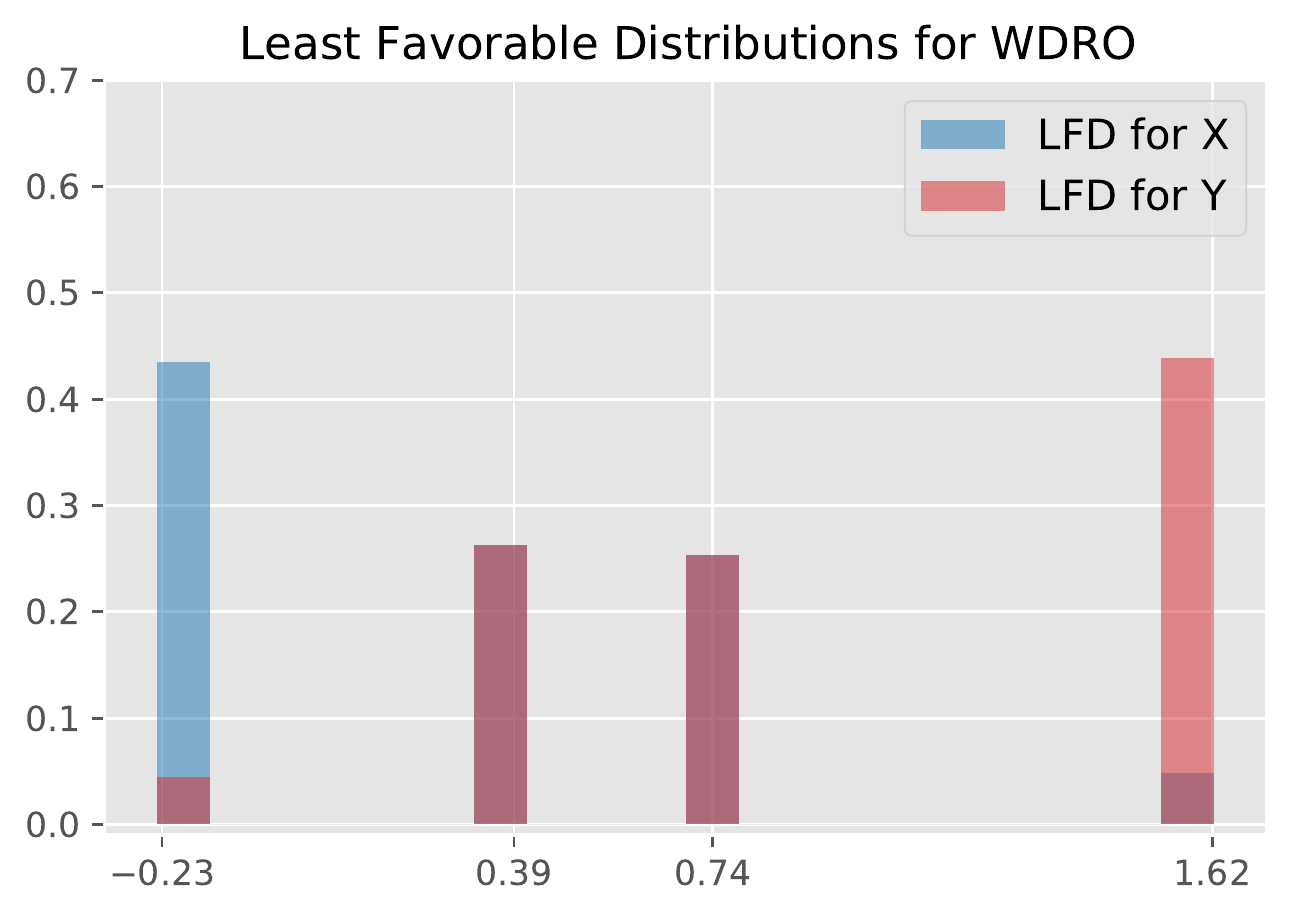}
    \includegraphics[width=0.24\textwidth]{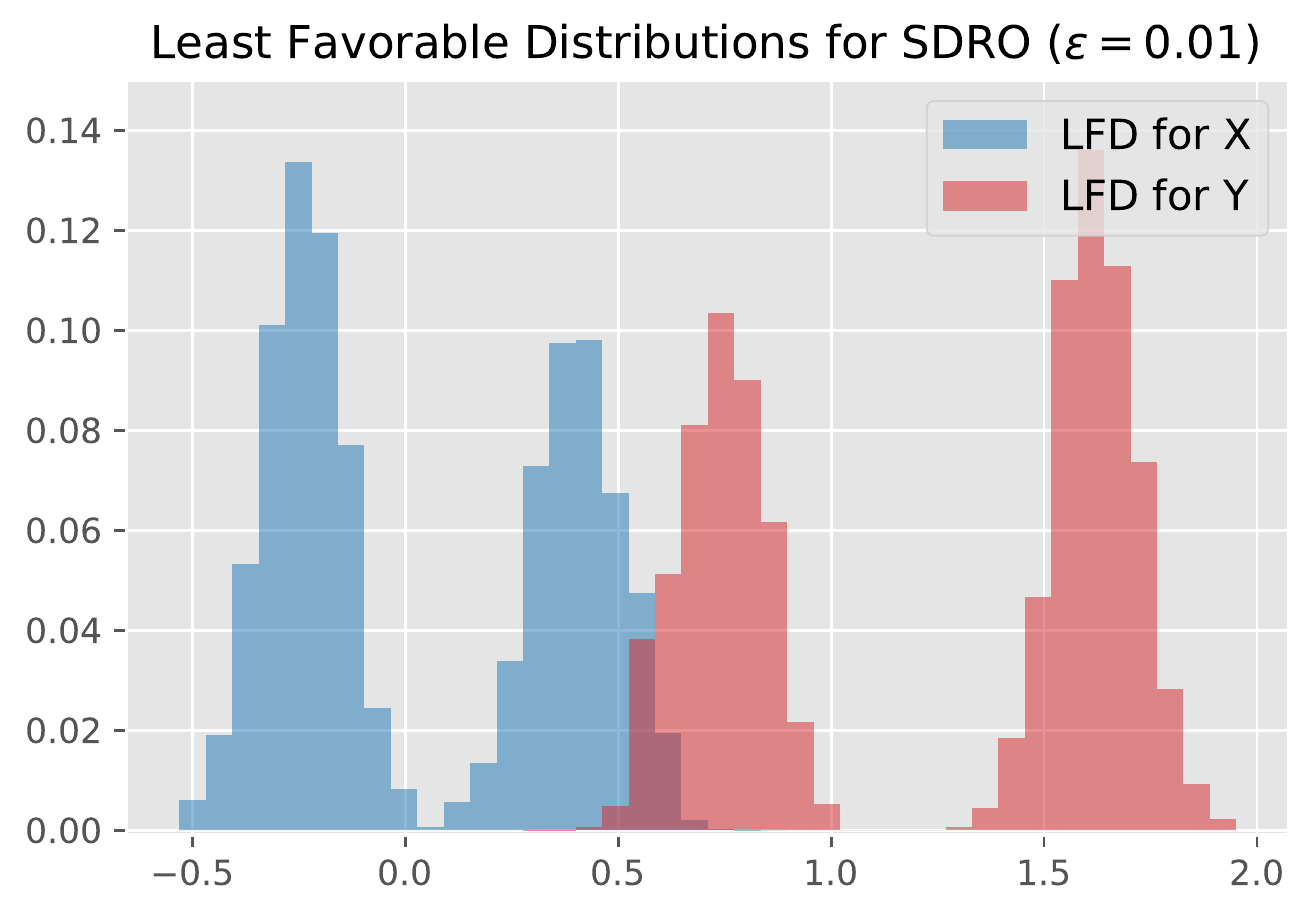}
    \includegraphics[width=0.24\textwidth]{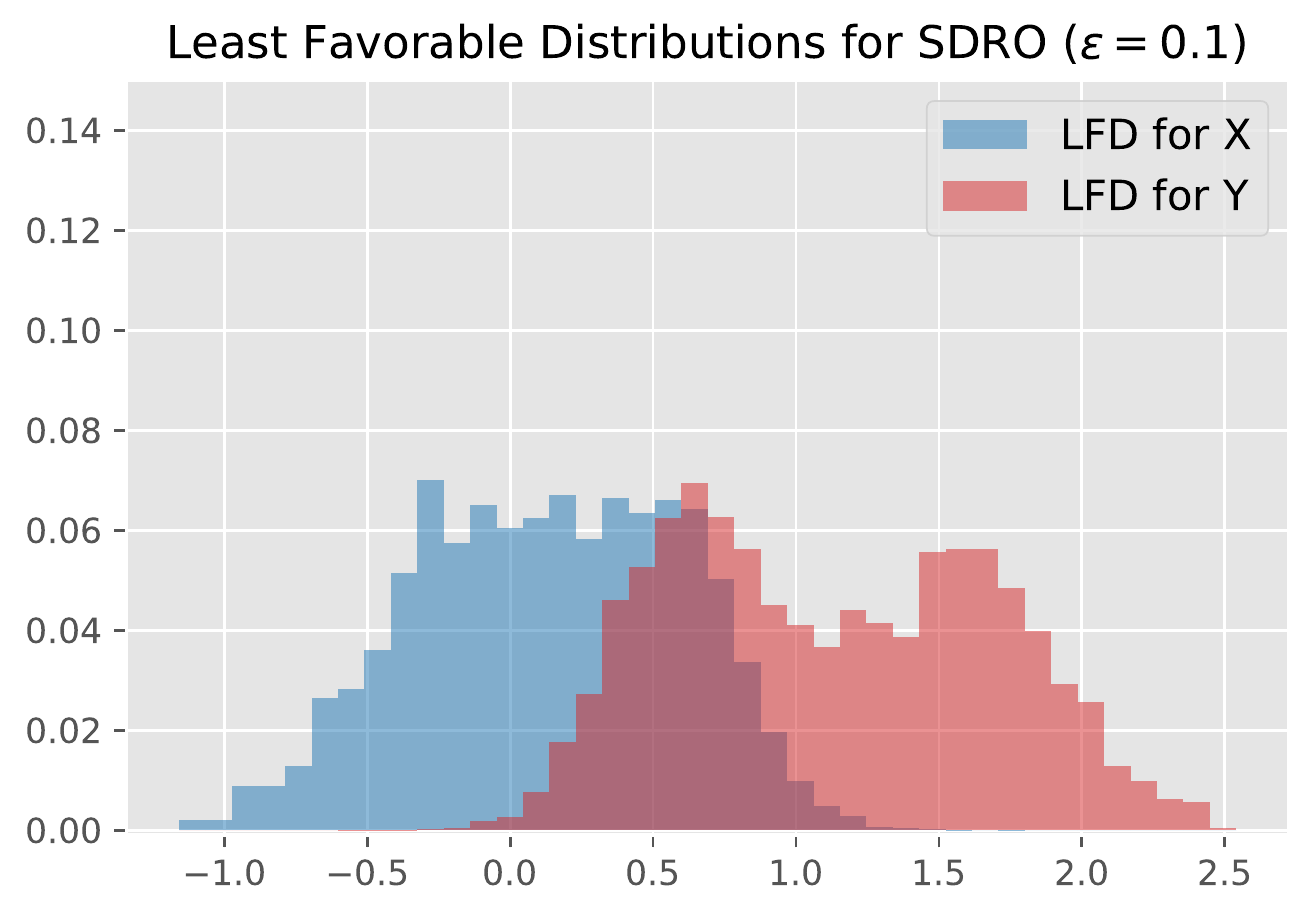}
    \includegraphics[width=0.24\textwidth]{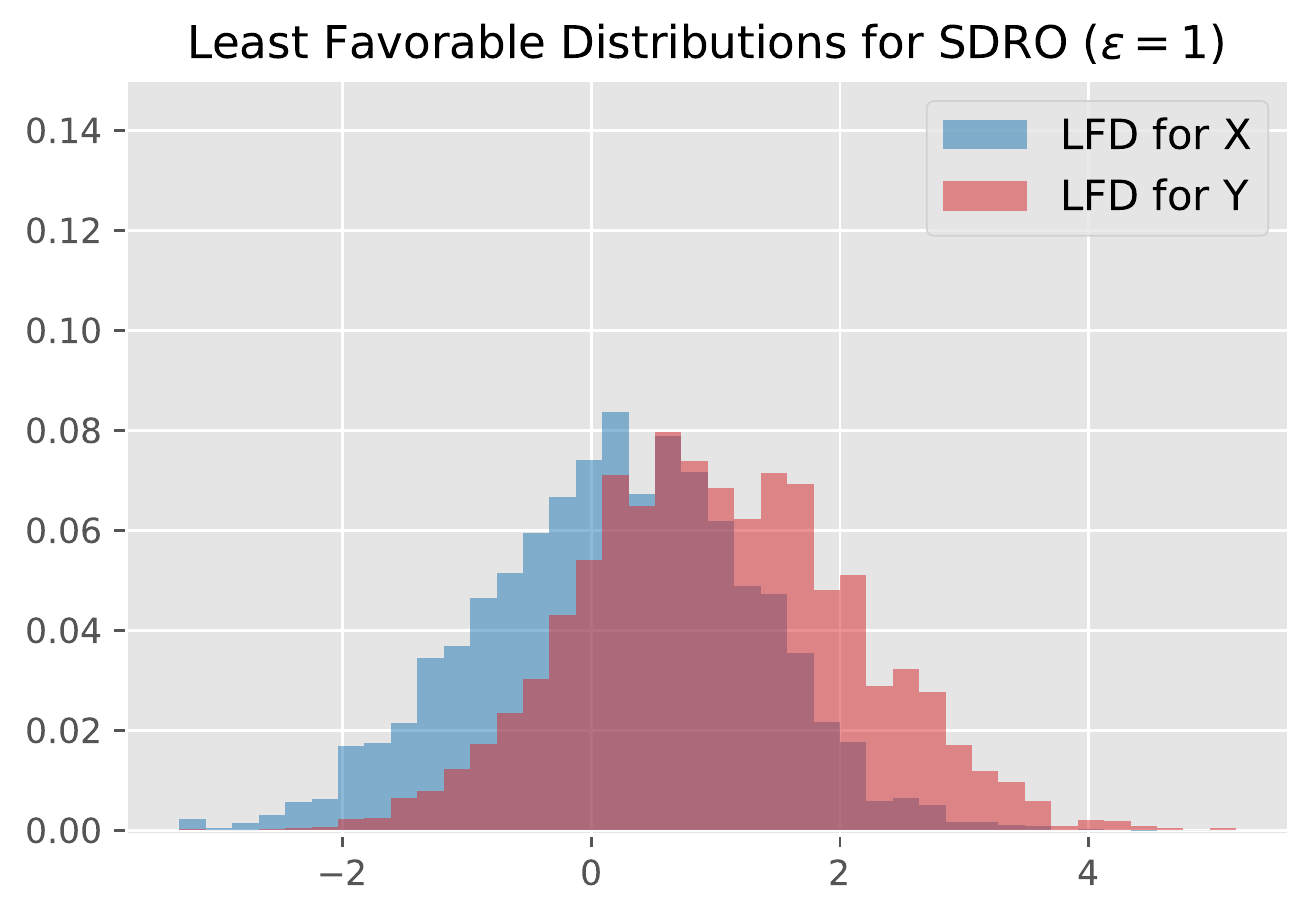}
    \caption{Plots of LFDs for WDRO test and SDRO test.}
    \label{fig:LFD}
\end{figure}
\fi

\if\submitversion2
\begin{figure}[H]
    \centering
    \includegraphics[width=0.4\textwidth]{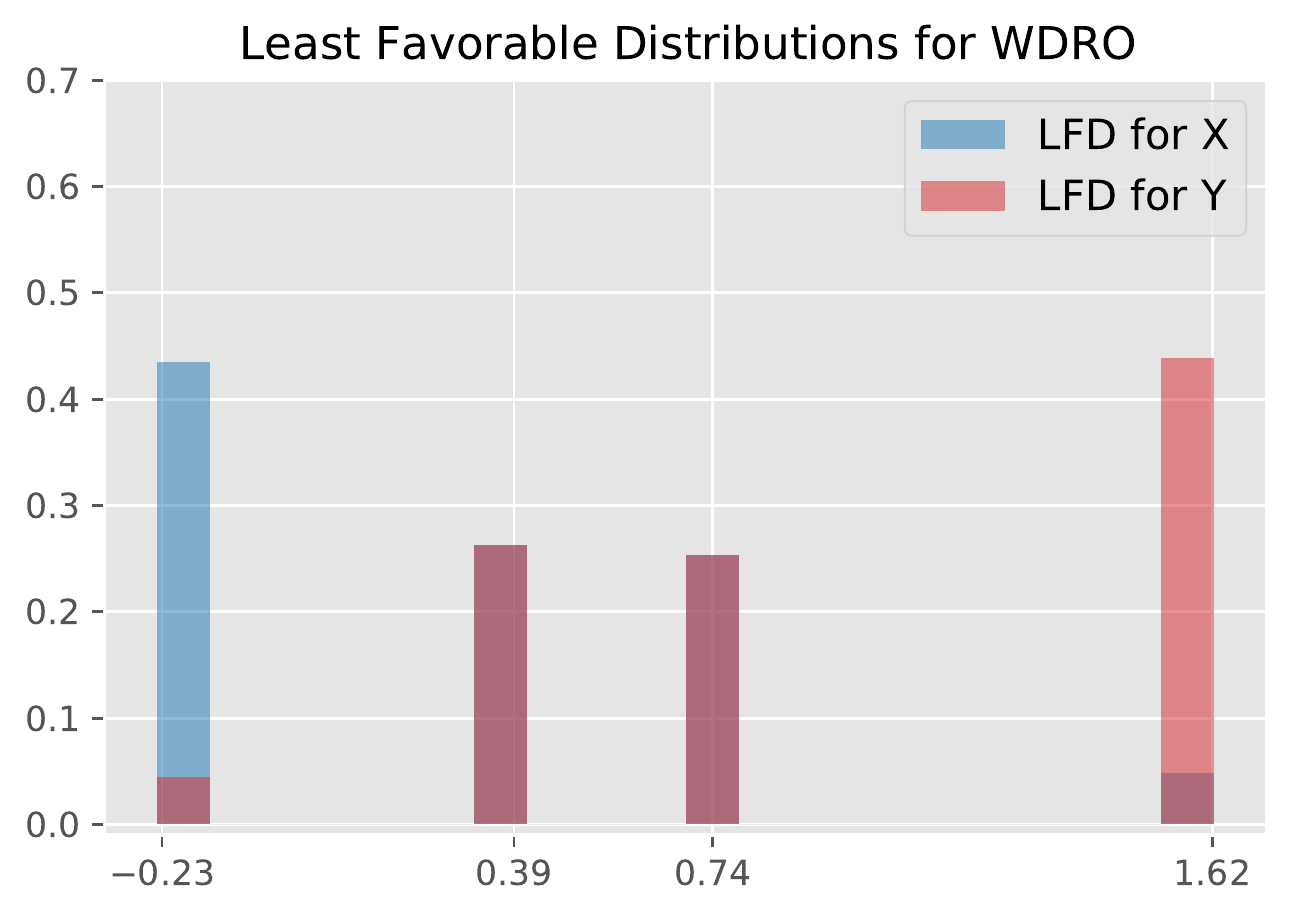}
    \includegraphics[width=0.4\textwidth]{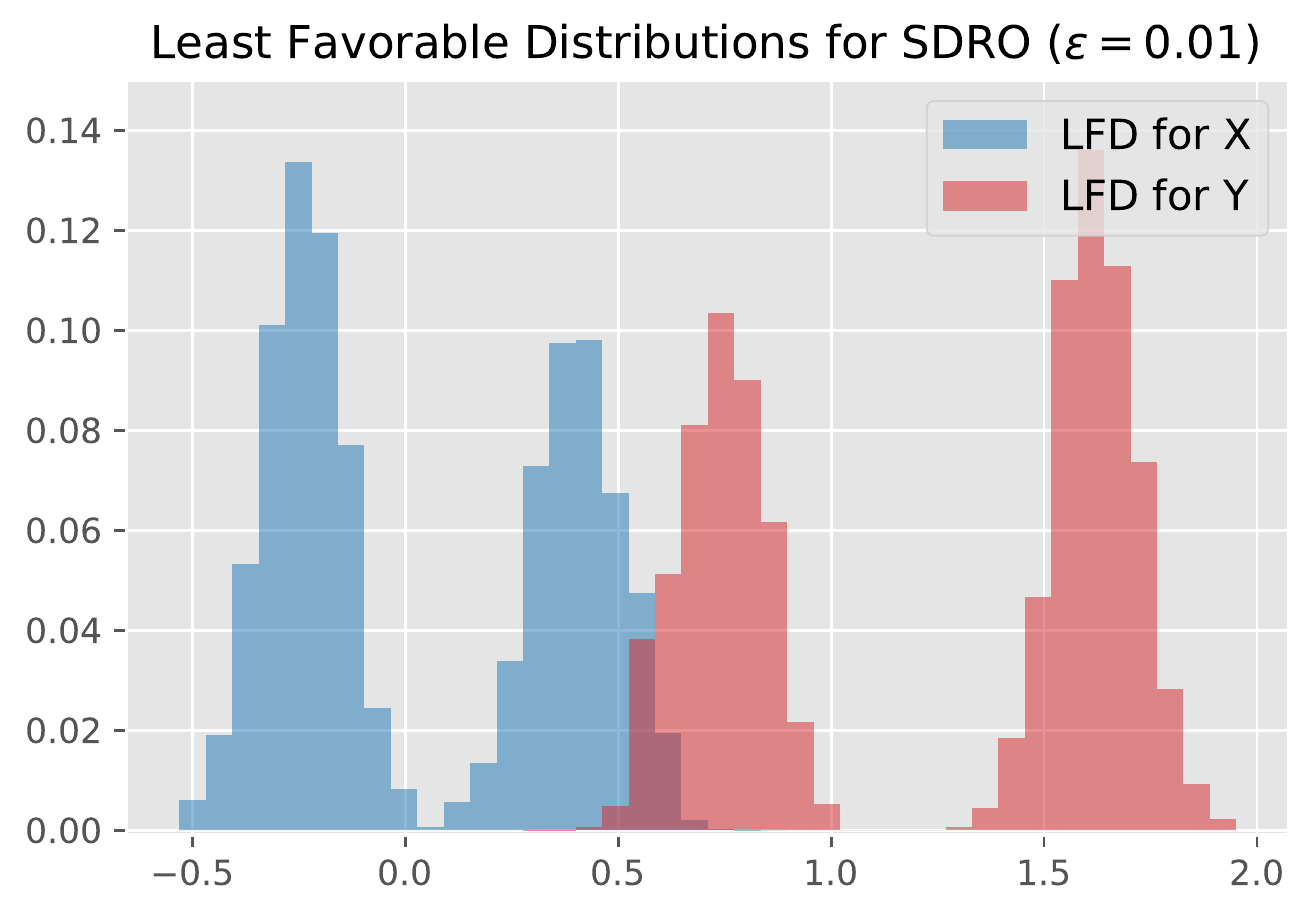}
    \includegraphics[width=0.4\textwidth]{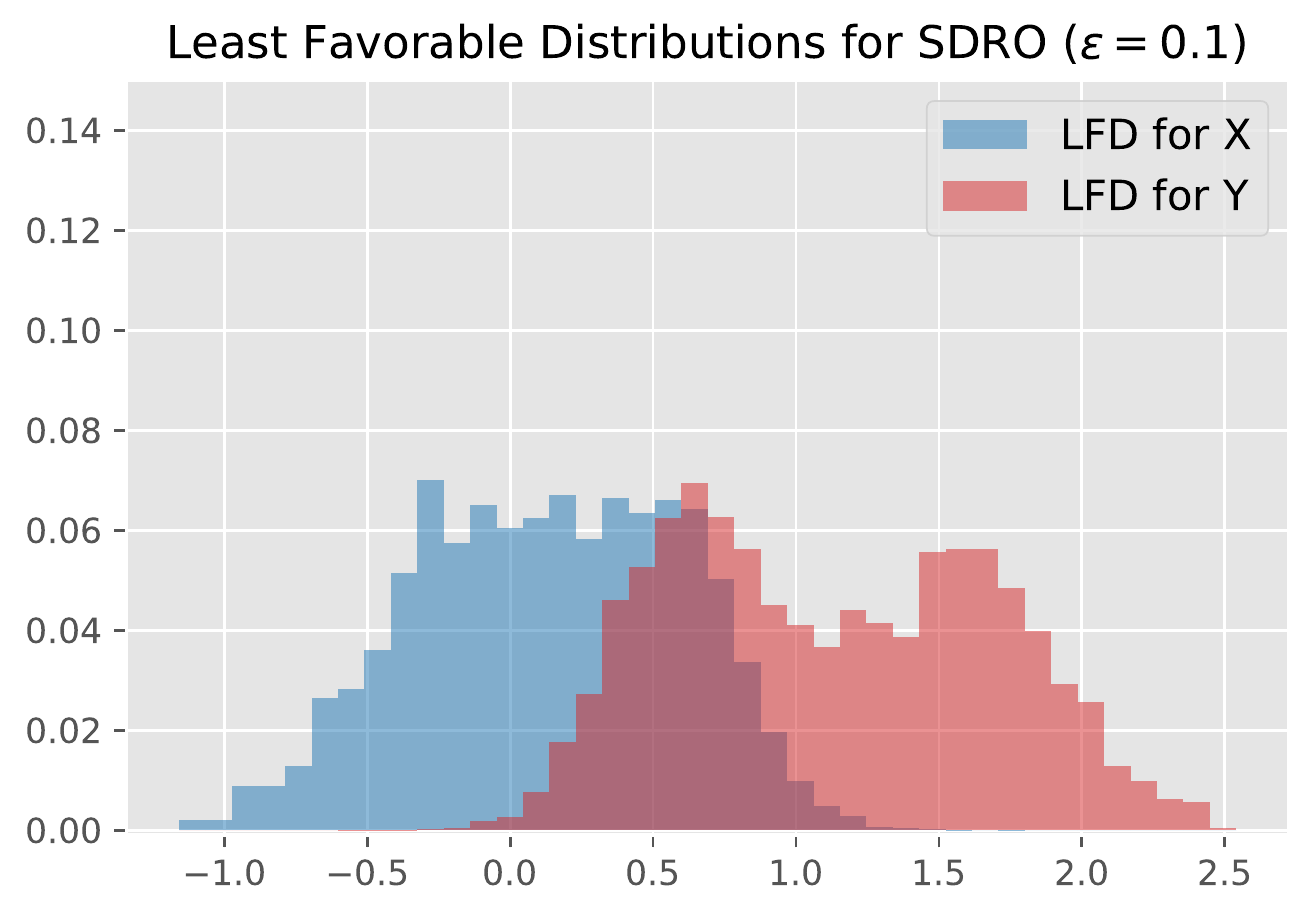}
    \includegraphics[width=0.4\textwidth]{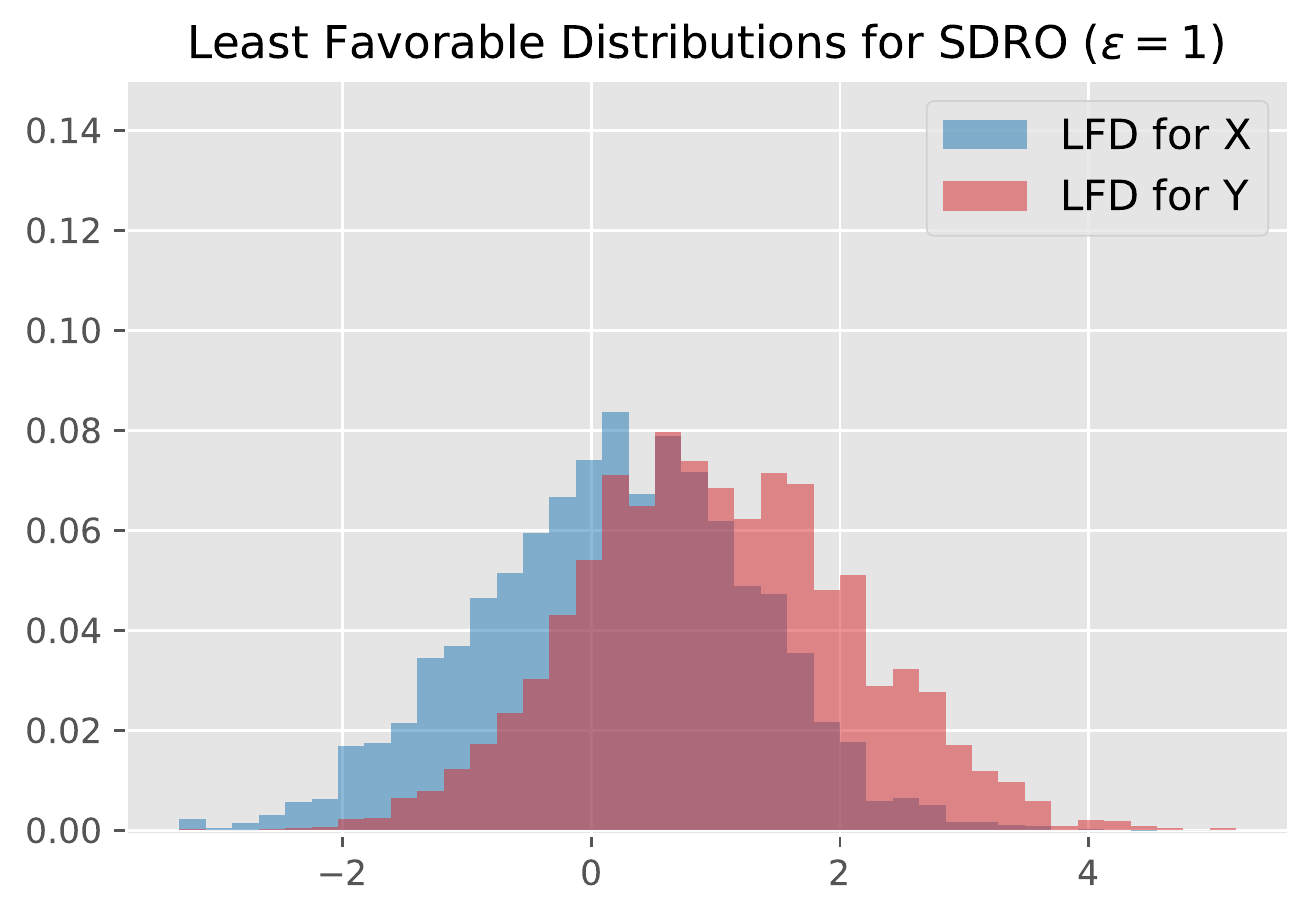}
    \caption{Plots of LFDs for WDRO test and SDRO test.}
    \label{fig:LFD}
\end{figure}
\fi

\end{document}